\documentclass[runningheads]{llncs}
\usepackage[caption=false]{subfig}
\usepackage{graphicx}
\usepackage{multirow}
\usepackage{hhline}
\usepackage{amsmath,amssymb}
\usepackage{bm}
\usepackage{color}
\usepackage{mdwlist}
\usepackage[width=122mm,left=12mm,paperwidth=146mm,height=193mm,top=12mm,paperheight=217mm]{geometry}
\begin{document}
\pagestyle{headings}
\mainmatter

\title{Improving Deep Neural Network with Multiple Parametric Exponential Linear Units} 

\author{Yang Li, Chunxiao Fan, Yong Li, Qiong Wu, Yue Ming}
\institute{Beijing University of Posts and Telecommunications}

\maketitle

\begin{abstract}
Activation function is crucial to the recent successes of deep neural networks. In this paper, we first propose a new activation function, Multiple Parametric Exponential Linear Units (MPELU), aiming to generalize and unify the rectified and exponential linear units. As the generalized form, MPELU shares the advantages of Parametric Rectified Linear Unit (PReLU) and Exponential Linear Unit (ELU), leading to better classification performance and convergence property. In addition, weight initialization is very important to train very deep networks. The existing methods laid a solid foundation for networks using rectified linear units but not for exponential linear units. This paper complements the current theory and extends it to the wider range.
Specifically, we put forward a way of initialization, enabling training of very deep networks using exponential linear units. Experiments demonstrate that the proposed initialization not only helps the training process but leads to better generalization performance. Finally, utilizing the proposed activation function and initialization, we present a deep MPELU residual architecture that achieves state-of-the-art performance on the CIFAR-10/100 datasets. The code is available at https://github.com/Coldmooon/Code-for-MPELU.
\end{abstract}

\section{Introduction}

Over the past few years, the landscape of computer vision has been noticeably changed from the engineered feature architecture to an end-to-end feature learning architecture, deep neural networks, by which many state-of-the-art work advanced the development of classical tasks such as object detection \cite{girshick2014rich}, semantic segmentation \cite{long2015fully}, and image retrieval \cite{li2015weakly}. Such a revolutionary change mainly results from several crucial elements, such as big datasets, high-performance hardware, new effective models, and regularization techniques. In this work, we focus on two notable elements, activation function and the corresponding initialization of network.

One of known activation functions is Rectified Linear Unit (ReLU) \cite{nair2010rectified,krizhevsky2012imagenet} which produced profound effect on the development of deep neural networks. ReLU is a piecewise-linear function that keeps positive inputs and outputs zero for negative inputs. Owing to this form, it can alleviate the problem of vanishing gradient, allowing the supervised training of much deeper neural networks. However, it experiences a potential disadvantage that units will never activate once gradients reach zero. Seeing this, Maas \emph{et al.} \cite{maas2013rectifier} presented Leaky ReLU (LReLU) where the negative part of activation function is replaced with a linear function. He \emph{et al.} \cite{He_2015_ICCV} further extended LReLU to a Parametric Rectified Linear Unit (PReLU) which can learn the parameters of the rectifiers, leading to higher classification accuracy with little overfitting risk. In addition, Clevert \emph{et al.} \cite{clevert2015fast} presented the Exponential Linear Unit (ELU), leading to faster learning and better generalization performance than the rectified unit family on deep networks. The above rectified and exponential linear units are commonly adopted by the recent deep learning architectures \cite{krizhevsky2012imagenet,simonyan2014very,Szegedy_2015_CVPR,he2015deep} to achieve good performance. However, there exists a gap of representation space between the two types of activation functions. For the negative part, ReLU or PReLU are able to represent the linear function family but not the non-linear one, while ELU is able to represent the non-linear function family but not the linear one. The representation gap to some extent undermines the representational power of those architectures using a particular activation function. In addition, ELU is at a potential disadvantage when used with Batch Normalization \cite{DBLP:conf/icml/IoffeS15}. Clevert \emph{et al.} \cite{clevert2015fast} showed that using Batch Normalization with ELU could harm the classification accuracy, which is also verified in our experiments.

This work is mainly motivated by PReLU and ELU. Firstly, we present a new Multiple Parametric Exponential Linear Unit (MPELU), a generalization of ELU, to bridge the gap. In particular, an extra learnable parameter, $\beta$, is introduced into the inputs of ELU to control the shape of negative part. By optimizing $\beta$ through stochastic gradient descent (SGD), MPELU is able to adaptively switch between the rectified and exponential linear units. Secondly, motivated by PReLU, we make the hyper-parameter $\alpha$ of ELU learnable to further improve its representational ability and tune the function shape. This design makes MPELU more flexible than its antecedents, ReLU, PReLU, and ELU that can be seen as special cases of MPELU. Therefore, through learning $\alpha$ and $\beta$, the linear and non-linear space of the negative part can be covered in a single activation function module, whereas its special existing cases do not have this property.

The introduction of learnable parameters into ELU may likely bring an additional benefit. This is inspired by the observation that Batch Normalization does not improve ELU networks but can improve ReLU and PReLU networks. To see this, MPELU can be inherently decomposed into a composition of PReLU and learnable ELU:
\begin{align}
\label{MPELU_decompose_no_BN}
     MPELU = \widetilde{ELU}[PReLU(x)],
\end{align}
where x is the inputs of activation function, and $\widetilde{ELU}$ denotes the ELU \cite{clevert2015fast} with a learnable parameter $\alpha$. Applying Batch Normalization to the inputs gives
\begin{align}
\label{MPELU_decompose}
     MPELU = \widetilde{ELU}\{ PReLU[ BN(x) ] \}.
\end{align}
As we can see, the outputs of Batch Normalization flow into PReLU before ELU, which can result in not only the improvement of the classification performance, but the alleviation of the potential problem of working with ELU. Eqn.~(\ref{MPELU_decompose}) suggests that MPELU is also able to share the advantages of PReLU and ELU simultaneously, for example, the superior learning behavior of ELU compared to ReLU and PReLU, as described in \cite{clevert2015fast}. Our experimental results on CIFAR-10 and ImageNet 2012 demonstrate that by introducing the learnable parameters, MPELU networks provide better classification performance and convergence property than its counterparts.

Because of the introduction of extra parameters, overfitting could be a concern. To address this, we adopt the same strategy as PReLU to reduce the overfitting risk. For each MPELU layer, $\alpha$ and $\beta$ are initialized as the channel-share version or the channel-wise version. Therefore, the increment of parameters of the entire network is at most twice the total number of channels, which is negligible compared to the number of weights. 

Although lots of activation functions, e.g., ELU \cite{clevert2015fast}, were proposed recently, few works determine a weight initialization for networks using them. Improper initialization often hampers the learning of very deep networks \cite{simonyan2014very}. Glorot \emph{et al.} \cite{glorot2010understanding} proposed an initialization scheme but only considered the linear activation functions. He \emph{et al.} \cite{He_2015_ICCV} derived an initialization method that considers the rectifier linear units (e.g., ReLU) but not makes allowance for the exponential linear units (e.g., ELU). Even though Clevert \emph{et al.} \cite{clevert2015fast} applied it to the networks using ELU, this lacks theoretical analysis. Furthermore, none of these works is suitable for non-convex activation functions. Observing this,
this paper presents a strategy of weight initialization, enabling the training of networks using exponential linear units including ELU and MPELU, and thus extends the current theory to the wider range. In particular, since MPELU is non-convex, the proposed initialization also applies to non-convex activation functions.

The main contributions of this work are:
\begin{enumerate*}
  \vspace{-10pt}
\item A new activation function MPELU that covers the solution space of both the rectified and exponential linear units.   
\item A technique of weight initialization, allowing the training of extremely deep networks using ELU and MPELU.
\item A simple architecture of ResNet with MPELU, achieving state-of-the-art results on the CIFAR \cite{krizhevsky2009learning} dataset with comparable time/memory complexity and parameters to the original versions \cite{he2015deep,He2016}.
\end{enumerate*}

The remainder of this paper is organized as follows. Sec.~\ref{section-2:related_work} reviews the related work. In Sec.~\ref{Section-3: the proposed methods}, we propose our activation function and initialization method. The experiments and analysis are given in Sec.~\ref{Section-4: MPELU_experiments} to show their effectiveness. Utilizing the proposed methods, Sec.~\ref{Section-5: Deep MPELU Residual Networks} presents a deep MPELU residual architecture to provide state-of-the-art performance on CIFAR-10/100. Finally, Sec.~\ref{Section-6: conclusion} concludes. To keep the paper at a reasonable length, the implementation details of our experiments are given in appendix.

\section{Related Work}
\label{section-2:related_work}

This paper mainly focuses on activation functions and the weight initialization of deep neural networks. Therefore, we review the related work in the two fields. Note that training very deep networks can also be realized by developing new architectures such as introducing skip connection as in \cite{NIPS2015_5850,he2015deep}, but this is beyond the scope of the paper.

\noindent \\
\textbf{Activation Functions.} Even though activation functions are an early invention, they were not formally defined until recently \cite{gulcehre2016noisy}. Activation functions allow deep neural networks to learn a complex non-linear transformation, which is crucial to the power of modeling. From the feature point of view, the outputs of activation functions can be used as high-level semantic representations (can also be obtained by subspace learning, e.g., \cite{li2015robust}) that are more robust to variance than low-level ones, which facilitates recognition tasks.

Among recent work is Rectified Linear Unit (ReLU) \cite{nair2010rectified,krizhevsky2012imagenet}, one of keys to the breakthrough of deep neural networks. ReLU keeps positive inputs unchanged and outputs zero for negative inputs, and therefore it can avoid the problem of vanishing gradients, enabling the training of much deeper supervised neural networks, whereas sigmoid nonlinearity can not. LReLU \cite{maas2013rectifier} was proposed that multiplies the negative inputs by a slope factor, aiming to avoid zero gradients in ReLU. According to \cite{maas2013rectifier}, LReLU provides comparable performance to ReLU and is sensitive to the value of the slope. He \emph{et al.} \cite{He_2015_ICCV} found that the cost function is differentiable with respect to the slope factor and therefore proposed optimizing the slope through SGD. This parametric rectified linear unit is named PReLU. Experiments showed that PReLU can improve the performance of convolutional neural networks with little overfitting risk. They also proved that PReLU has the ability of pushing off-diagonal blocks of FIM closer to zero, which enables faster convergence than ReLU. None of the above activation functions can learn the non-convex functions since their essence of convex function. To address this, Jin \emph{et al.} \cite{Jin2016aaai} proposed a S-shaped rectified linear activation unit (SReLU) to learn both convex and non-convex functions, which is inspired by the Webner-Fechner law and the Stevens law. In addition to the above rectified linear units, Clevert \emph{et al.} \cite{clevert2015fast} presented a novel form of activation function, Exponential Linear Unit (ELU). ELU is similar to sigmoid for negative inputs and has the same form as ReLU for positive inputs. It has been proved that ELU is able to bring the gradient closer to the unit natural gradient, which accelerates learning speed and leads to higher performance. When used with Batch Normalization \cite{DBLP:conf/icml/IoffeS15}, ELU tends to expose an unexpected degradation problem. In this case, ELU has a negligible impact on the generalization capability and classification performance. In addition to the above deterministic activation functions, there is another random version. Recently, Xu \emph{et al.} \cite{xu2015empirical} proposed a randomized leaky rectified linear unit, RReLU. RReLU also has negative values which is helpful to avoid zero gradients. The difference is that the slope of RReLU is not fixed or learnable but randomized. Through this strategy, RReLU is able to reduce the overfitting risk to some extent. However, Xu \emph{et al.} only verified RReLU on small datasets, like CIFAR-10/100. How RReLU performs on large datasets such as ImageNet is still needed to be explored. 

\noindent \\
\textbf{Initialization.} Initialization of parameters is very important especially for deep networks and the case of large learning rate. If not initialized properly, it may be very hard to converge through SGD. Many efforts have concentrated on this subject. Hinton \emph{et al.} \cite{hinton2006fast} introduced a learning algorithm that utilizes layer-wise unsupervised pre-training to initialize all layers. Before this, there is no suitable algorithms for training deep fully-connected architectures. Shortly after, Bengio \emph{et al.} \cite{bengio2007greedy} studied the pre-training strategy and conducted a series of experiments to substantiate and verify it. Erhan \emph{et al.} \cite{erhan2009difficulty} further performed a number of experiments to confirm and clarify the procedure, showing that it can initialize the starting point in parameter space in a better basin of attraction than picking starting parameters at random. During the development of deep learning, another important work is ReLU \cite{nair2010rectified} which addresses the problem of vanishing gradients. With ReLU, deep networks are able to converge even randomly initialized from a Gaussian distribution. Krizhevsky \emph{et al.} \cite{krizhevsky2012imagenet} applied ReLU to supervised convolutional neural networks with random initialization and won the ILSVRC 2012 challenge. Since that, deeper and deeper networks have been proposed, leading to a sequence of improvements in computer vision. However, Simonyan \emph{et al.} \cite{simonyan2014very} showed that deep networks still face the optimization problem once the number of layers reaches some value (e.g., 11 layers). This phenomenon is also mentioned in \cite{glorot2010understanding,Szegedy_2015_CVPR,He_2015_ICCV,NIPS2015_5850}. Glorot \emph{et al.} \cite{glorot2010understanding} proposed a method to initialize weights according to the size of a layer. This strategy holds under the assumption of linear activation functions, which works well in many cases but not holds for rectified linear units (e.g., ReLU and PReLU). He \emph{et al.} \cite{He_2015_ICCV} extended this method to the case of rectified linear units and proposed a new initialization strategy usually MSRA filler which has shown great help for training very deep networks. Nevertheless, for exponential linear units, there is currently no appropriate strategy to initialize weights. Observing this, we generalize the MSRA filler to a new initialization for deep networks using exponential linear units (e.g., ELU and MPELU) based on the first-order Taylor expansion of MPELU at zero. 

\begin{figure}[t]
     \centering
     \subfloat[][shapes of activation functions]{
     \includegraphics[width=0.4\textwidth]{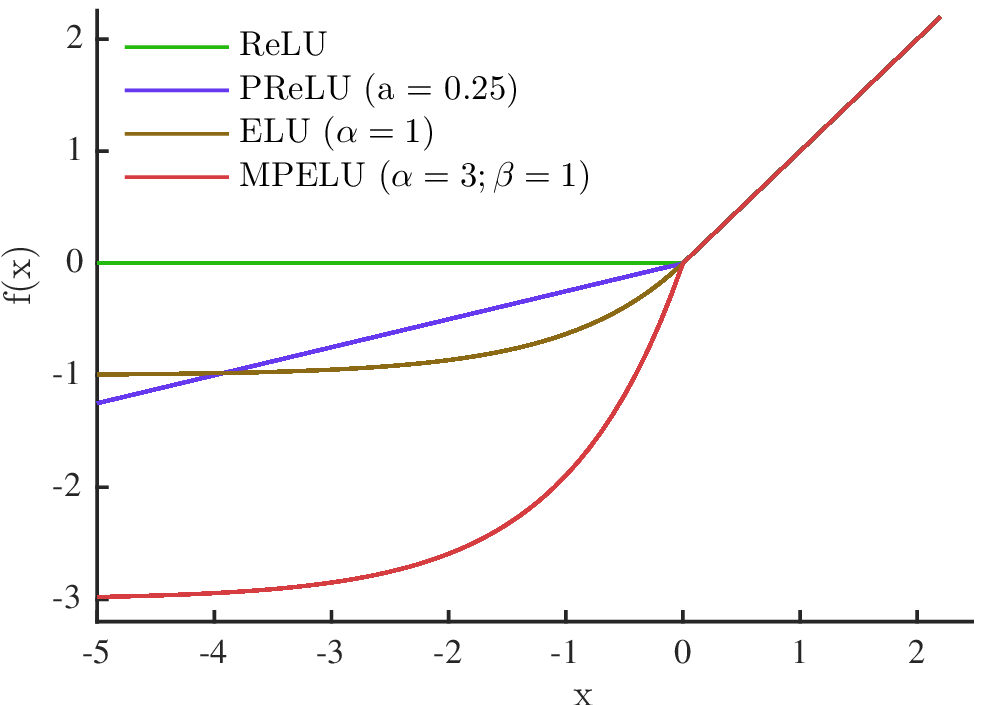}\label{fig:1-a}}
     ~~~~~~~~
     \subfloat[][other activation functions are special cases of MPELU]{
     \includegraphics[width=0.4\textwidth]{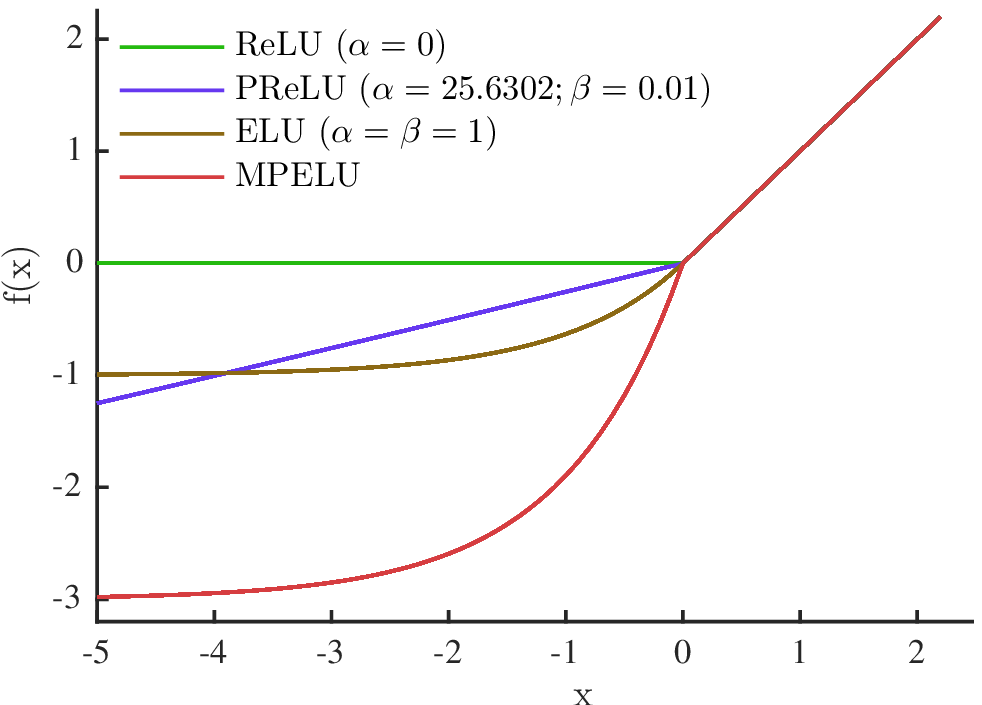}\label{fig1-b}}
     \caption{The graphical depiction of activation functions. (a) shapes of activation functions. $a$ of
		PReLU is initialized with 0.25. The hyper-parameter $\alpha$ of ELU is 1. $\alpha$ and $\beta$ of MPELU are initialized with 3 and 1, respectively. (b) other activation functions are special cases of MPELU. With $\alpha$ = 0, MPELU is reduced to ReLU. If $\alpha$ = 25.6302 and
		$\beta$ =0.01, MPELU approximates to PReLU; When $\alpha$, $\beta$ = 1, MPELU becomes
		ELU}
     \label{fig:1}
\end{figure}

\section{The Proposed Activation Function and Weight Initialization}
\label{Section-3: the proposed methods}

This section first presents the Multiple Parametric Exponential Linear Unit (MPELU), then derives the weight initialization for networks using exponential linear units.

\subsection{Multiple Parametric Exponential Linear Unit}

PReLU and ELU have limited but complementary representational power for their negative parts. This work proposes a general form of activation function that unifying the existing ReLU, LReLU, PReLU, and ELU.

\noindent \\
\textbf{Forward Pass.} Formally, the definition of MPELU is:
\begin{align}
\label{MPELU_forward}
	f(y_i)=\left\{\begin{matrix}
	y_i & if & y_i>0\\ 
	\alpha_c (e^{\beta_c y_i}-1) & if & y_i\leqslant0 &.
	\end{matrix}\right.
\end{align}
Here, $\beta$ is constrained to be greater than zero, and $i$ is the index of input $y$ corresponding to the $c_{th}$ ($c \in \{1, ... ,M\}$) $\alpha$ and $\beta$. Following PReLU, $\alpha_c$ and $\beta_c$ can be channel-wise ($M =$ the number of feature maps) or channel-shared ($M = 1$) learnable parameters, which control the value to and at which MPELU saturates respectively. Fig.~1(a) shows the shapes of four activation functions.

By adjusting $\beta_c$, MPELU can switch between the rectified and exponential linear units. To be specific, if $\beta_c$ is set to a small number, for example, 0.01, the negative part of MPELU approximates to a linear function. In this case, MPELU becomes the Parametric Rectified Linear Unit (PReLU). On the other side, if $\beta_c$ takes a large value, for example, 1.0, the negative part of MPELU is a non-linear function, making MPELU turn back into the exponential linear units. 

Introducing $\alpha_c$ helps further control the form of MPELU, as shown in Fig.~1(b). If $\alpha_c$ and $\beta_c$ are set to 1, MPELU reduces to ELU. Decreasing $\beta_c$ in this case lets MPELU go to LReLU. Finally, MPELU is exactly equivalent to ReLU when $\alpha_c = 0$.

From the above analysis, it is easy to see that the flexible form of MPELU makes it cover the solution space of its special cases, and therefore grants it more powerful representation. We will show that ResNet \cite{he2015deep,He2016} could gain significant improvement merely by tuning the usage of activation functions, that is, from ReLU to MPELU.

Another benefit of MPELU is fast learning. Eqn.~(\ref{MPELU_decompose}) suggests that MPELU could potentially share the properties of PReLU and ELU. Thus, as an exponential linear unit, MPELU exhibits the same learning behavior as ELU. Readers are referred to \cite{clevert2015fast} for more details.

\noindent \\
\textbf{Backward Pass.} Since MPELU is differentiable almost everywhere, deep networks with MPELU can be trained end-to-end. We use chain rule to derive the update formulations of $\alpha_c$ and $\beta_c$: 
\begin{align}
\label{Eqn: backward pass of MPELU 1}
top'&=f(y_i)+\alpha_c\\
\frac{\partial f(y_i)}{\partial \alpha_c}&=\left\{\begin{matrix}
0 \ \ \ & if & y_i>0\\ 
\label{Eqn: backward pass of MPELU 2}
e^{\beta_c y_i}-1 \ \ \ & if & y_i\leqslant0
\end{matrix}\right.\\
\label{Eqn: backward pass of MPELU 3}
\frac{\partial f(y_i)}{\partial \beta_c}&=\left\{\begin{matrix}
0 \ \ \ & if & y_i>0\\ 
y_i*top' \ \ \ & if & y_i\leqslant0
\end{matrix}\right.\\
\label{Eqn: backward pass of MPELU 4}
\frac{\partial f(y_i)}{\partial y_i}&=\left\{\begin{matrix}
1 \ \ \ & if & y_i>0\\ 
\beta_c*top' \ \ \ & if & y_i\leqslant0 &.
\end{matrix}\right.
\end{align}

Note that $\frac{\partial f(y_i)}{\partial \alpha_c}$ and $\frac{\partial f(y_i)}{\partial \beta_c}$ are the gradients of activation function with respect to $\alpha_c$ and $\beta_c$ for a single unit. When computing the gradients of loss function for the entire layer, the gradients of $\alpha_c$ and $\beta_c$ should be: 
\begin{align}
	\frac{\partial L}{\partial \alpha_c}&=\sum_{y_i}\frac{\partial L}{\partial f(y_i)}*\left\{\begin{matrix}
	0 \ \ \ & if & y_i>0\\ 
	e^{\beta_c y_i}-1 \ \ \ & if & y_i\leqslant0 
	\end{matrix}\right.\\
	\frac{\partial L}{\partial \beta_c}&=\sum_{y_i}\frac{\partial L}{\partial f(y_i)}*\left\{\begin{matrix}
	0 \ \ \ & if & y_i>0\\ 
	y_i*top'_i \ \ \ & if & y_i\leqslant0 &,
	\end{matrix}\right.
\end{align}
where $\Sigma$ sums over all the positions corresponding to $\alpha_c$ and $\beta_c$. Throughout this paper, we employ the channel-wised version for all the experiments. By this strategy, the increment of parameters of the entire network is at most twice the total number of channels, which is negligible compared to the number of weights. We show in Sec.~\ref{Section-5: Deep MPELU Residual Networks} that the model size of the proposed MPELU ResNet architectures can be comparable to (or even less than) that of ReLU architectures. 

For the actual running time, MPELU is roughly comparable to PReLU if we carefully optimize the codes. This will be analyzed in Section \ref{Section: experiments on ImagenNet}.

Initializing $\alpha$ and $\beta$ with different values has small but non-negligible impact on classification accuracy. We recommend using $\alpha = 1 \ or \ 0.25$ and $\beta = 1$ as the initial values, and five times the base learning rate for both of them. Moreover, we highlight that it is important to use weight decay  ($l_2$ regularization) on both $\alpha$ and $\beta$, which is opposite the case of rectified linear units such as PReLU \cite{He_2015_ICCV} and SReLU \cite{Jin2016aaai}. 

\subsection{The Proposed Weight Initialization for Networks with MPELU}
\label{Section: weight initialization}

The previous works \cite{hinton2006fast,bengio2007greedy,glorot2010understanding,He_2015_ICCV} have laid a solid foundation for the initialization of deep neural networks. This paper complements the current theory and extends it to the wider range.

\noindent \\
\textbf{Briefly Review of MSRA filler.} MSRA filler contains two cases of initialization, the forward propagation case and the backward propagation case. He \emph{et al.} \cite{He_2015_ICCV} proved that both cases are able to properly scale the backward signal. Therefore, it is sufficient to only investigate the forward propagation case.

For the $l_{th}$ convolutional layer, a pixel in the output channel is expressed as:
\begin{align}
	y_l = \bm{w_l} * \bm{x_l} + b_l,
\end{align}
where $y_l$ is a random variable, \bm{$w_l$} and \bm{$x_l$} are random vectors and independent of each other, and $b_l$ is initialized with zero. The goal is to explore the relationship between the variance of $y_{l-1}$ and the variance of $y_l$.
\begin{align}
	Var(y_l) = Var(\bm{w_l} \bm{x_l} + b_l) = Var(\bm{w_l} \bm{x_l}) = k_l^{2} c_l Var(w_l x_l ),
\label{dyl}
\end{align}
where $k_l$ is the kernel size and $c_l$ is the number of input channels. Here, both $w_l$ and $x_l$ are random variables. Eqn.~(\ref{dyl}) holds under the assumption that the elements in \bm{$w_l$} and \bm{$x_l$} are independent and identically distributed respectively. 
Usually, weights of deep network are initialized with zero mean, and Eqn.~(\ref{dyl}) becomes:
\begin{align}
	Var(y_l) = k_l^2 c_l Var(w_l)E(x_l^2).
	\label{Eqn: variance of y_l}
\end{align}
Next, we need to find the relationship between $E(x_l^2)$ and $Var(y_l-1)$. Note that there exists an activation function between $x_l$ and $y_{l-1}$,
\begin{align}
	x_l = f(y_{l-1}).
\end{align}
For different activation functions $f$, we may derive different relationships, and thus different initialization methods. Specifically, for symmetric activation functions, the sigmoid non-linearity, Glorot \emph{et al.} \cite{glorot2010understanding} assumed they are linear at the initialization and therefore proposed the Xavier method. For rectified linear units, ReLU and PReLU, He \emph{et al.} \cite{He_2015_ICCV} removed the linear assumption and extended the Xavier method to the MSRA filler. In the next section, we further extend the MSRA filler to a more general form by taking the first-order Taylor expansion of MPELU at zero and clipping the results to its linear part.

\noindent \\
\textbf{The Proposed Initialization.} This section mainly follows the derivation in \cite{glorot2010understanding,He_2015_ICCV}. Since ELU is a special case of MPELU, we focus on MPELU. As we can see from Eqn.~(\ref{MPELU_forward}), it is very difficult to obtain the exact relationship between $E(x_l^2)$ and $Var(y_{l-1}$). Instead, we use its Taylor series at zero. For the negative part, MPELU can be expressed as:
\begin{align}
\alpha (e^{\beta y}-1)=\alpha \beta y+\frac{1}{2}\alpha(\beta y)^{2}+\frac{1}{3!}\alpha(\beta y)^{3}+... \ .
\label{taylor_express}
\end{align}
Then, the left side of Eqn.~(\ref{taylor_express}) is approximated by its Taylor polynomial of degree 1.  
\begin{align}
	\alpha(e^{\beta y}-1) =\alpha \beta y+R_n(y) \approx \alpha \beta y
	\label{assumption_linearity}
\end{align}
Eqn.~(\ref{assumption_linearity}) introduces the linear approximation only for the negative regime. We call this semi-linear assumption with which we have: 
\begin{align}
x_l &\approx \max (0,y_{l-1})+ \min (0,\alpha \beta y_{l-1}) \\
E(x_l^2) &= \int_{-\alpha}^{\infty }x_l^2 p(x_l) dx_l \approx \frac{1}{2} (1 + \alpha_{l-1}^2 \beta_{l-1}^2) E(y_{l-1}^2),
\end{align}
where, $p(x)$ is the probability density function. Following \cite{glorot2010understanding,He_2015_ICCV}, if $w_{l-1}$ having a symmetric distribution with zero mean, it is also the case for $y_{l-1}$. Then,  
\begin{align}
E(x_l^2) \approx \frac{1}{2} (1 + \alpha_{l-1}^2 \beta_{l-1}^2) Var(y_{l-1}).
\label{Exl_Dyl-1}
\end{align}
By Eqn.~(\ref{Exl_Dyl-1}) and (\ref{Eqn: variance of y_l}), we obtain:
\begin{align}
	Var(y_l) \approx \frac{1}{2}k_l^2 c_l (1 + \alpha_{l-1}^2 \beta_{l-1}^2) Var(w_l) Var(y_{l-1}).
\end{align}
Through this, it is easy to derive the relationship between $y_{l-1}$ and $y_1$: 
\begin{align}
Var(y_l) \approx Var(y_1)\prod_{i=2}^{l}\frac{1}{2}k_i^2 c_i (1 + \alpha_i^2 \beta_i^2) Var(w_i).
\end{align}
Following \cite{glorot2010understanding,He_2015_ICCV}, to keep the signals of the forward and backward pass flowing correctly, we expect that $Var(y_1)$ is equal to $Var(y_l)$, which leads to: 
\begin{align}
	\frac{1}{2}k_i^2 c_i (1 + \alpha_i^2 \beta_i^2) Var(w_i)=1, \forall i.
\end{align}
Therefore, for each layer in deep networks using MPELU, we can initialize weights from a Gaussian distribution 
\begin{align}
	\left ( 0, \sqrt{\frac{2}{k_i^2 c_i (1 + \alpha_i^2 \beta_i^2)}}\ \right ),
\label{taylor_result}
\end{align}
where $i$ is the index of layer. Eqn.~(\ref{taylor_result}) applies to deep networks using the rectified or exponential linear units.
Note that when $\alpha = 1$ and $\beta = 1$, Eqn.~(\ref{taylor_result}) becomes the initialization for ELU networks. When $\alpha = 0$, Eqn.~(\ref{taylor_result}) corresponds to the initialization for ReLU networks. Furthermore, when $\alpha = 0.25$ and $\beta = 1$, Eqn.~(\ref{taylor_result}) can be used to initialize PReLU networks. From this point of view, MSRA filler is a special case of the proposed initialization.

\noindent \\
\textbf{Comparison with Xavier, MSRA, and LSUV.} Xavier method is designed for symmetric activation functions with the hypothesis of linearity, and MSRA filler only applies to the rectified linear units (ReLU and PReLU), while the proposed method addresses the initialization for both rectified and exponential linear units. Recently, Mishkin \emph{et al.} \cite{mishkin2015all} proposed the LSUV initialization that is data-driven and thus avoids solving the relationship between $E(x_l^2)$ and $Var(y_l-1)$, but Eqn.~(\ref{taylor_result}) is an analytic solution for ELU and MPELU and therefore runs faster than LSUV.

\section{Experiment}
\label{Section-4: MPELU_experiments}

This section explores the usage of MPELU on a number of architectures. In Sec.~\ref{Section: experiments on cifar10}, we begin with experiments with Network in Network (NIN) \cite{2013arXiv1312.4400L} on CIFAR-10, showing the benefit of introducing learnable parameters into ELU. Sec.~\ref{Section: experiments on ImagenNet} further substantiates this benefit in deeper networks and on the larger dataset, ImageNet 2012. Finally, Sec.~\ref{sect:initialization experiments} verifies the proposed initialization with a very deep network on ImageNet, showing the ability of training very deep ELU/MPELU networks. In Sec.~\ref{Section: experiments on cifar10} and Sec.~\ref{sect:initialization experiments}, we also provide the convergence analysis, showing that MPELU, like ELU, possesses the superior convergence property to ReLU and PReLU.

\subsection{Experiments with NIN on CIFAR-10}
\label{Section: experiments on cifar10}

This section conducts the experiments of Network in Network with different activation functions on the CIFAR-10 dataset. The goal is to investigate the benefits of introducing learnable parameters into ELU. 

This architecture has nine convolutional layers including six ones with $1\times1$ kernel size and no Fully Connected (FC) layers, which is easy to train and sufficient for a comprehensive evaluation of effectiveness of learnable parameters. The implementation details are given in appendix.

\renewcommand{\arraystretch}{0.7}
\setlength{\tabcolsep}{4pt}
\begin{table}[t]
	\begin{center}
		\caption{Test error rate (\%) of classification on the CIFAR-10. $\alpha$ and $\beta$ in MPELU are initialized with 1 or 0.25, and they are updated by SGD without weight decay. As in \cite{NIPS2015_5850,he2015deep} the best (mean $\pm$ std) results are reported by five runs for each network}
		\label{table:cifar10}
		\begin{tabular}{llll}
			\hline\noalign{\smallskip}
			NIN & parameter(s) & CIFAR-10 & CIFAR-10 (augmented)\\
			\noalign{\smallskip}
			\hline
			\noalign{\smallskip}
			ReLU \cite{2013arXiv1312.4400L} & - & 10.41 & 8.81\\
			PReLU & $\alpha = 0.25$ & \textbf{9.02 (9.19 $\pm$ 0.15)} & \textbf{7.28 (7.49 $\pm$ 0.14)}\\
			ELU & $\alpha = 1$ & 9.39 (9.63 $\pm$ 0.23) & 7.77 (7.83 $\pm$ 0.05)\\
			MPELU & $\alpha = 1$; $\beta = 1$ & 9.06 (9.19 $\pm$ 0.11) & 7.37 (7.57 $\pm$ 0.16) \\
			MPELU & $\alpha = 0.25$; $\beta = 1$ & 9.10 (9.27 $\pm$ 0.12) & 7.30 (7.52 $\pm$ 0.18)
			\\
			\hline
		\end{tabular}	
	\end{center}
\end{table}

For fair comparison, we train networks using ReLU, PReLU, ELU, and MPELU with the same settings from scratch. Tab.~\ref{table:cifar10} shows that MPELU consistently outperforms ELU (e.g., 9.06\% vs. 9.39\% test error rate without data augmentation, and 7.30\% vs 7.77\% test error rate with data augmentation). This improvement over ELU is completely from $\alpha$ and $\beta$, verifying the benefit from the learnable parameters.

\setlength{\tabcolsep}{1.4pt}
\begin{figure}[t]
     \centering
     \subfloat[][training loss]{
     \label{fig: nin training loss on cifar}
     
     \includegraphics[width=1\textwidth]{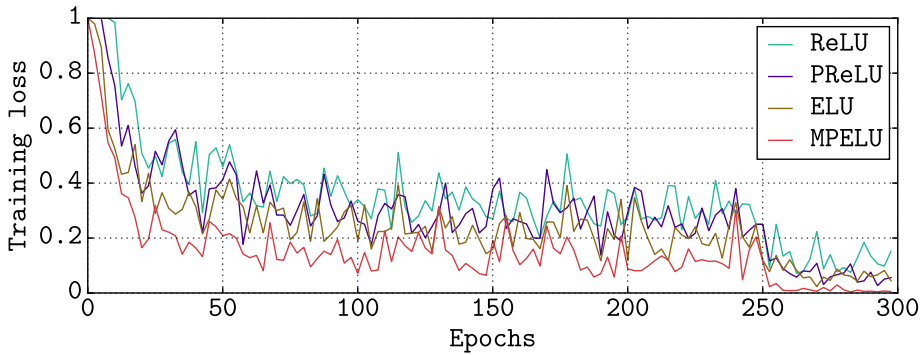}\label{fig:2-a}}\\
     \subfloat[][test error rate]{
     \label{fig: nin test error on cifar}
     \includegraphics[width=1\textwidth]{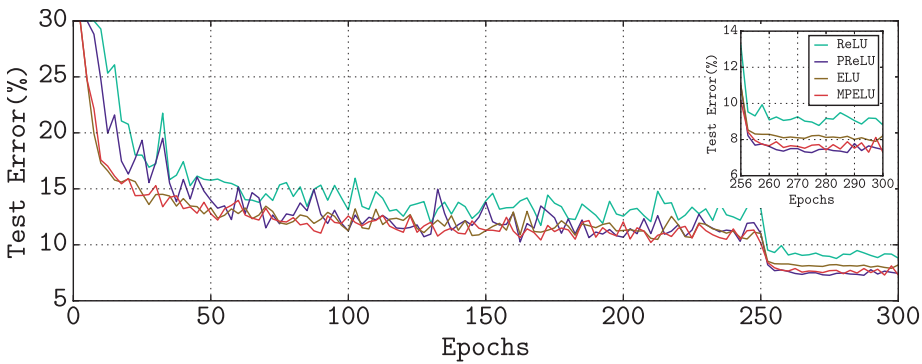}\label{fig:2-b}}
     \caption{Comparison of convergence on CIFAR-10. All the models learn very quickly on this small dataset, and so we adopt the evaluation method similar to \cite{clevert2015fast} according to which the number of iterations used to reach 15\% test error is measured. 
     (a) indicates that MPELU can reduce the loss earlier. (b) shows that MPELU reaches the 15\% error after 9k iterations, while ReLU and PReLU need 25k and 15k iterations to reach the same error rate}
     \label{fig:2}
\end{figure}

Some interesting phenomenon can be observed in Tab.~\ref{table:cifar10} and Fig.~\ref{fig:2}. Firstly, Tab.~\ref{table:cifar10} shows that MPELU ($\alpha = \beta = 0.25$) performs like PReLU (a negligible difference of 0.03\% mean test error when using data augmentation). Secondly, Fig.~\ref{fig:2}(a)(b) show that its learning curves are closer to ELU's, suggesting a potential superior learning behavior compared to the rectified linear units, as described in \cite{clevert2015fast}. Note that all the models learn very quickly on this small dataset and reach the same test error rate (15\%) within 25k iterations, which makes it very hard to compare the learning speed. To deal with this, we adopt the similar evaluation criterion to \cite{clevert2015fast}, that is, the iteration to reach the 15\% test error rate. Fig.~\ref{fig:2}(b) shows that MPELU starts reducing the error (also the loss) earlier and reaches the 15\% error after 9k iterations, while ReLU and PReLU need 25k and 15k iterations to reach the same error rate, respectively. The above better performance arises from the combining advantages of PReLU and ELU, as suggested in Eqn.~(\ref{MPELU_decompose_no_BN}). 

It is also worth noting that MPELU achieves a comparable performance to PReLU with a bit more parameters. This is not caused by overfitting since ELU performs much worse than PReLU and MPELU. The underlying reason is still unclear and will be studied in the future. Even though MPELU is a bit less effective than PReLU in this shallower architecture, we will show that MPELU outperforms PReLU in deeper architectures.

\subsection{Experiments on ImageNet}
\label{Section: experiments on ImagenNet}

\renewcommand{\arraystretch}{0.7}
\setlength{\tabcolsep}{4pt}
\begin{table}[t]
\begin{center}
\caption{Top-1 error rate (single-view test) on the validation set of ImageNet 2012 with data augmentation. The comparison is under the same initial values of $\alpha$. $\beta$ in MPELU is initialized with 1 for all cases. $\alpha$ and $\beta$ in MPELU are updated by SGD with/without weight decay. MPELU outperforms its counterparts consistently and obtains the overall best result} 
\label{table:modele15_imagenet}
\begin{tabular}{|c|c|c|c|c|c|c|c|c|c|}
\hline
$\alpha$ & $\beta$ & \multicolumn{4}{c|}{Gaussian initialization} & \multicolumn{2}{c|}{MSRA} & \multicolumn{2}{c|}{our initialization} \\
\hline
\multicolumn{2}{|l|}{$\beta$ for MPELU} & ReLU & PReLU & ELU & MPELU & ReLU & PReLU & ELU & MPELU \\
\hline
0/0 & 1/0 & 37.66 & - & - & 39.40 & 37.45 & - & - & - \\
0/1 & 1/0 & -     & - & - & 37.92 & -     & - & - & - \\
0/1 & 1/1 & -     & - & - & \textbf{37.61} & - & - & - & \textbf{37.41} \\
\hline 
0.25/0 & 1/0 & - & 39.48 & - & 40.94 & - & 38.72 & - & 39.46 \\
0.25/1 & 1/1 & - & 39.53 & - & \textbf{37.81} & - & 38.57 & - & \textbf{37.47} \\
\hline 
1/0 & 1/0 & - & - & 40.36 & 39.53 & - & - & 39.83 & 38.42 \\
1/1 & 1/1 & - & - & -     & \textbf{38.04} & - & - & - & \textbf{\color{blue}{37.33}} \\
\hline 
\multicolumn{6}{l}{$\alpha$, $\beta$: initial value / weight decay multiplier } \\
\end{tabular}
\end{center}
\end{table}

This section evaluates MPELU on the ImageNet 2012 classification task. ImageNet 2012 contains about 1.28 million training examples, 50k validation examples, and 100k test examples which belong to 1000 classes. This enables us to utilize a deeper network with little overfitting risk. Therefore, we build a 15-layer network modified from the model-E in \cite{He_2015_CVPR}. The models evaluated in this section are trained on the training set and tested on the validation set. \\
\noindent \\
\textbf{Network Structure.} Based on the model-E, we add one more convolutional layer, insert Batch Normalization \cite{DBLP:conf/icml/IoffeS15} immediately before activation functions, and remove dropout \cite{JMLR:v15:srivastava14a} layers. Following \cite{He_2015_CVPR,krizhevsky2012imagenet,sppnet}, the networks are divided into three stages by max-pooling layers. The first stage contains only one convolutional layer with a kernel size of $7\times7$ pixels and 64 filters. The second stage consists of four convolutional layers with the kernel size of $2\times2$ pixels and 128 filters. We set stride and pad accordingly so as to maintain the feature map size of $36\times36$ pixels. The third stage consists of seven convolutional layers with kernel size of $2\times2$ pixels and 256 filters. In the third stage, the feature map size is reduced to $18\times18$ pixels. The next layer is spp \cite{sppnet} which is followed by two 4096-d FC layers, one 1000-d FC layer, and one softmax successively. The networks are initialized through three methods which are Gaussian distribution with zero mean and 0.01 standard deviation, MSRA filler \cite{He_2015_ICCV}, and the proposed initialization (see Sec. \ref{Section: weight initialization}). The bias terms are initialized with 0 as usual. $\alpha$ and $\beta$ in MPELU are initialized with varying values and updated by SGD with weight decay. Other implementation details are given in appendix.

For fair comparison, the participants are evaluated under the same initial values of $\alpha$, and Tab. \ref{table:modele15_imagenet} lists the results. For clarity, the results that outperform others are marked in boldface and the overall best result is marked in blue.\\

\noindent 
\textbf{Gaussian Initialization.} When compared to ELU, all the MPELU layers are initialized with $\alpha = \beta = 1$. As we can see, the MPELU network outperforms the ELU network by 0.83\% top-1 error rate. If weight decay is used, it can significantly outperform the ELU network by 2.32\%. Since the only difference between them lies in the activation function, this improvement over ELU indeed demonstrates the advantage of the learnable parameters, $\alpha$ and $\beta$. 

For further examining MPELU, we also compared it with PReLU. In this case, $\alpha$ in MPELU are initialized with 0.25. Tab.~\ref{table:modele15_imagenet} shows that the MPELU network achieves the top-1 error rate 40.94\%, which is worse than 39.48\% provided by the PReLU network. Nevertheless, using weight decay considerably improves the performance of the MPELU network by 3.13\%, reducing the top-1 error rate to 37.81\% which is better than that of the PReLU network by 1.72\%. 

\noindent \\
\textbf{Other Initialization Methods.} Experiments are also conducted with other initialization methods (see Tab.~\ref{table:modele15_imagenet}). The experimental results are in line with the Gaussian initialization case. MPELU surpasses all the counterparts. The overall best top-1 error rate 37.33\% achieved by MPELU is significantly lower than those achieved by PReLU and ELU. It is interesting to see that the MPELU networks initialized from the proposed method consistently outperform those initialized from Gaussian method, demonstrating that our initialization can lead to better generalization capability, which is also verified in Sec.~\ref{sect:initialization experiments}.

Note that MPELU only provides slight improvement over ReLU, and using weight decay in MPELU tends to decrease the top-1 test error in all three cases. This result is not caused, however, by overfitting, since adding more layers (more parameters) to the 15-layer network leads to lower test error, as shown in Sec.~\ref{sect:initialization experiments}. A possible reason is that using weight decay tends to push $\alpha$ and $\beta$ to zero, resulting in smaller scale activations or sparser representations, like ReLU, that are more likely to be linearly separable in a high-dimensional space \cite{glorot2011deep}. Another explanation may come from the sparse feature selection \cite{li2014clustering}.

To provide an empirical interpretation, we performed four extra experiments using LReLU with different slopes, and gradually decreased the scale of activations. All the five models (ReLU and LReLU A-D) have the same number of parameters, which eliminates the influence of overfitting. The only difference among them is the scale of the negative activations. A noticeable trend is illustrated in Tab.~\ref{table:4-leaky_relu_experiments}. The top-1/top-5 test error decreases with the slope, which explains why using weight decay to MPELU leads to better results and why ReLU performs better than PReLU and ELU. Nevertheless, this phenomenon is not observed in Sec.~\ref{Section: experiments on cifar10}, which might be due to that small scale or sparsity is less important for the shallower architecture (The ReLU NIN performs worst).

\setlength{\tabcolsep}{4pt}
\begin{table}[t]
\begin{center}
\caption{Classification comparison among different slopes on the ImageNet validation set. The trend is that the performance increases with the decrease of slope}
\label{table:4-leaky_relu_experiments}
\begin{tabular}{llcc}
\hline\noalign{\smallskip}
15-layer network with & slope & top-1 error rate(\%) & top-5 error rate(\%) \\
\noalign{\smallskip}
\hline
\noalign{\smallskip}
ReLU  & $a = 0$  & 37.66 & 15.98 \\
LReLU (A) & $a = 0.1$ & 37.92 & 16.26 \\
LReLU (B) & $a = 0.25$ & 38.54 & 16.65 \\
LReLU (C) & $a = 0.5$ & 42.76 & 20.18 \\
LReLU (D) & $a = 1$ & 60.27 & 36.60
\\
\hline
\end{tabular}
\end{center}
\end{table}
\setlength{\tabcolsep}{1.4pt}

\noindent \\
\textbf{Convergence Comparison.} Since Batch Normalization has a great influence on the convergence of networks, we leave the comparison of convergence among activation functions to Sec.~\ref{sect:initialization experiments}.

\noindent \\
\textbf{Running Time.} The running time refers to the time consumption of performing an iteration with batch size 64 during training. Essentially, the computational cost of MPELU is greater than its counterparts. But this problem can be properly addressed by carefully engineered implementation (e.g., faster exponential functions). In our Caffe \cite{Jia:2014:CCA:2647868.2654889} implementation, the backward pass utilizes the outputs of the forward pass, as shown in Eqn.~(\ref{Eqn: backward pass of MPELU 1})(\ref{Eqn: backward pass of MPELU 3})(\ref{Eqn: backward pass of MPELU 4}), which saves a lot of computation. Furthermore, the gradients of parameters and inputs can be computed together for each loop. Consequently, the real running time of MPELU can be only slightly slower than that of PReLU, as summarized in Tab.~\ref{table:runing_time}.

\renewcommand{\arraystretch}{1}
\setlength{\tabcolsep}{4pt}
\begin{table}[t]
	\begin{center}
		\caption{The running time (seconds/iteration) of ReLU, PReLU, ELU, and MPELU based on Caffe implementation. The experiments are performed on a NVIDIA Titan X GPU. The running time below is the mean value of 600k iterations}
		\label{table:runing_time}
		\begin{tabular}{lllll}
			\hline\noalign{\smallskip}
			  & ReLU & PReLU & ELU & MPELU \\
			\noalign{\smallskip}
			\hline
			\noalign{\smallskip}
			running time & 0.2310 & 0.2417 & 0.2299 & 0.2441 
			\\
			\hline
		\end{tabular}
	\end{center}
	\vspace{-20pt}
\end{table}
\setlength{\tabcolsep}{1.4pt}

\subsection{Experiments of Initialization}
\label{sect:initialization experiments}

This section conducts experiments on ImageNet 2012. The task is to examine whether the proposed initialization is able to help with convergence of very deep networks using exponential linear units. To this end, we add extra 15 convolutional layers to the network in Sec.~\ref{Section: experiments on ImagenNet}, resulting in a 30-layer network that suffices for investigating the effect of the initialization. Note that the network is similar to the 30-layer ReLU network in \cite{He_2015_ICCV} but differs from it in several aspects such as batch size, pad, and feature map size.

Since BN has a great influence on the convergence of deep networks, it is nature to take it into account. Following \cite{DBLP:conf/icml/IoffeS15}, we remove dropout layers when using BN. Finally, four methods are compared: the baseline Gaussian initialization, our initialization, BN + Gaussian initialization, and BN + our initialization. $\alpha$ and $\beta$ in MPELU are initialized with 1 and updated by SGD without weight decay, with other settings identical to Sec.~\ref{Section: experiments on ImagenNet}.

\setlength{\tabcolsep}{4pt}
\begin{table}[t]
\begin{center}
\caption{Comparison of initialization. The top-1 test error (\%) on the validation set of ImageNet 2012 is reported. The 30-layer ELU and MPELU networks with Gaussian method totally stop learning. On the contrary, the proposed method makes them converge, verifying the effectiveness of Eqn.~(\ref{taylor_result}). When BN is used, the performance can still be boosted by the proposed method. Note that the results, 44.28\% and 42.96\%, achieved by the 30-layer MPELU networks with BN are considerably lower than those, 39.53\% and 38.42\%, achieved by the 15-layer counterparts, suggesting the emergence of the degradation problem \cite{he2015deep} }
\label{Comparison of initialization}
\scalebox{0.9}{
\begin{tabular}{|l|c|c|c|c|} \hline
\multirow{2}{*}{initialization methods}& \multicolumn{2}{|c|}{30-layer networks} & \multicolumn{2}{|c|}{15-layer networks}\\ \cline{2-5}
                  & ELU      & MPELU                        & ELU            & MPELU          \\ \hline
Gaussian          & $\times$ & $\times$                     & -              & -              \\ \hline
ours              & 37.08    & \textbf{\color{blue}{36.49}} & -              & -              \\ \hline
Gaussian + BN     & -        & 44.28                        & 40.36          & 39.53          \\ \hline
ours + BN         & -        & \textbf{42.96}               & \textbf{39.83} & \textbf{38.42} \\ \hline
\multicolumn{5}{l}{$\times$: fails to converge} \\
\end{tabular}}
\end{center}
\end{table}
\setlength{\tabcolsep}{1.4pt}

\setlength{\tabcolsep}{4pt}
\begin{table}[t]
\begin{center}
\caption{Comparison between LSUV and ours through the 15-layer networks. Although the improvement over LSUV is slight, but still consistent} 
\label{table:comparison_for_15_lsuv}
\begin{tabular}{|c|c|c|c|c|c|c|}
\hline
15 layers & \multicolumn{5}{c|}{MPELU} & ELU \\
\hline
$\alpha$, $\beta$ & 0/1, 1/1 & 0.25/0, 1/0 & 0.25/1, 1/1 & 1/1, 1/1 & 1/0, 1/0 & 1/0, 1/0 \\
\hline
LSUV \cite{mishkin2015all} & 37.72 & 39.93 & 37.67 & 37.62 & 38.57 & 39.85 \\ \hline
ours & \textbf{37.41} & \textbf{39.46} & \textbf{37.47} & \textbf{37.33} & \textbf{38.42} & \textbf{39.83} \\ 

\hline 
\multicolumn{7}{l}{$\alpha$, $\beta$: initial value / weight decay multiplier}
\end{tabular}
\end{center}
\vspace{-10pt}
\end{table}

\noindent \\
\textbf{Comparison to Gaussian.} Tab.~\ref{Comparison of initialization} shows that the Gaussian initialization fails to train the 30-layer ELU/MPELU networks, while our method can help learn, which justifies the effectiveness of Eqn.~(\ref{taylor_result}). Furthermore, the 37.08\%/36.49\% top-1 test error rates achieved by the 30-layer ELU/MPELU networks are obviously lower than those achieved by 15-layer counterparts, meaning that the proposed method indeed addresses the diminishing gradients caused by improper initialization of very deep networks, hence makes them enjoy the benefit from the increase of depth. When BN is adopted, the proposed method reduces the error consistently compared to the Gaussian initialization, showing its benefit to the generalization capability. In addition, MPELU networks always perform better than ELU networks, and obtains the overall best result, 36.49\% top-1 test error rate, demonstrating the benefit of introducing learnable parameters. The above results indicate that although Eqn.~(\ref{taylor_result}) derives from a first-order Taylor approximation of Eqn.~(\ref{taylor_express}), it indeed works rather well in practice. 

\noindent \\
\textbf{Comparison to LSUV.} Mishkin \emph{et al.} \cite{mishkin2015all} verified LSUV in the 22-layer GoogLeNet \cite{Szegedy_2015_CVPR} using ReLU. To examine LSUV in deeper networks with exponential linear units, we build another 52-layer ELU network and initialize the 30- and 52-layer ELU networks with LSUV. Without BN, LSUV makes both ELU networks explode within only several iterations, while our method can make them converge. More experiments are also conducted through the 15-layer networks from Sec.~\ref{Section: experiments on ImagenNet} and the results are given in Tab.~\ref{table:comparison_for_15_lsuv}. The proposed initialization leads to marginal, but consistent, decrease in top-1 test error. In addition, Eqn.~(\ref{taylor_result}) is an analytic solution, while LSUV is a data-driven method, meaning that the proposed method runs faster than LSUV.

\noindent \\
\textbf{Degradation Analysis.} It should be noted in Tab.~\ref{Comparison of initialization} that while the 30-layer network without BN obtains the overall best result, the 30-layer networks with BN perform considerably worse than the 15-layer counterparts. To explain this, we analyze their learning behaviors.

\setlength{\tabcolsep}{1.4pt}
\begin{figure}[t]
     \centering
     \subfloat[][training loss (end)]{
     \includegraphics[width=\textwidth]{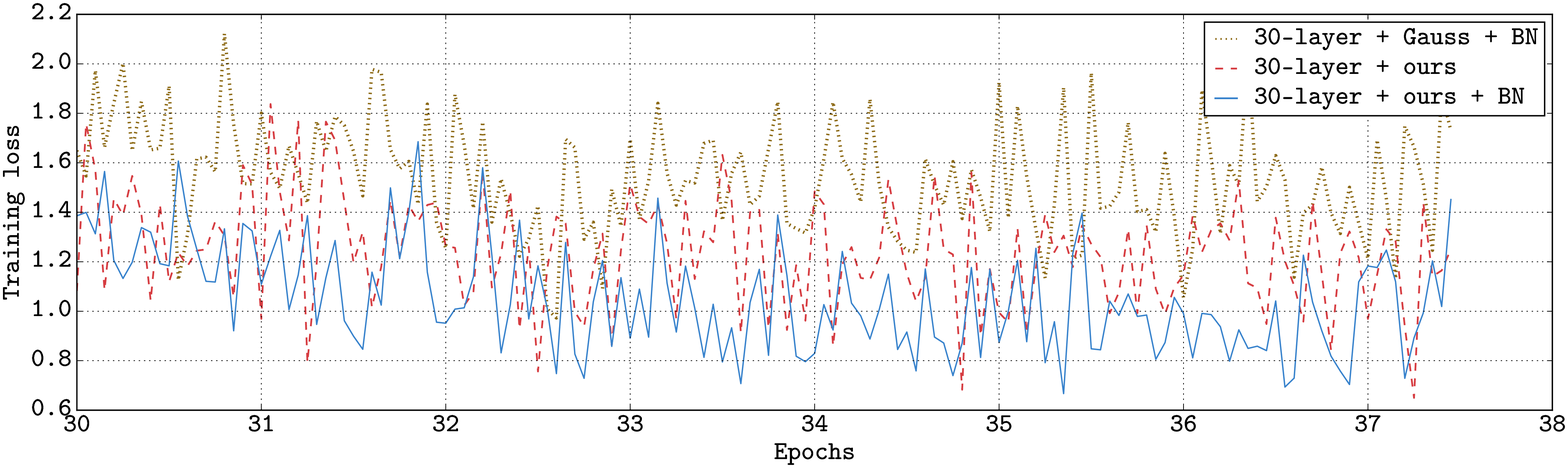}}
     \\
     \subfloat[][training error]{
     \includegraphics[width=0.5\textwidth]{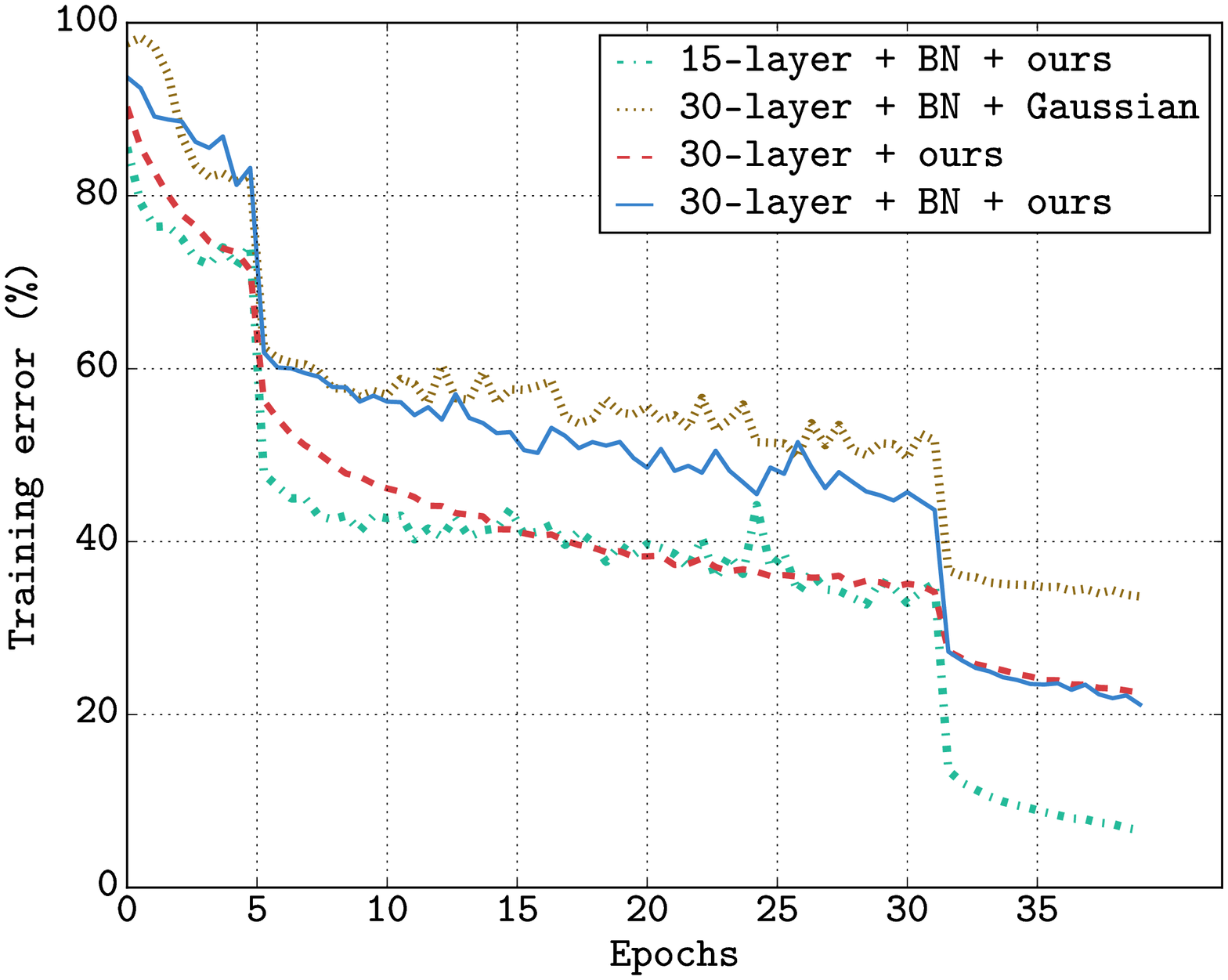}}
     \subfloat[][test error]{
     \includegraphics[width=0.5\textwidth]{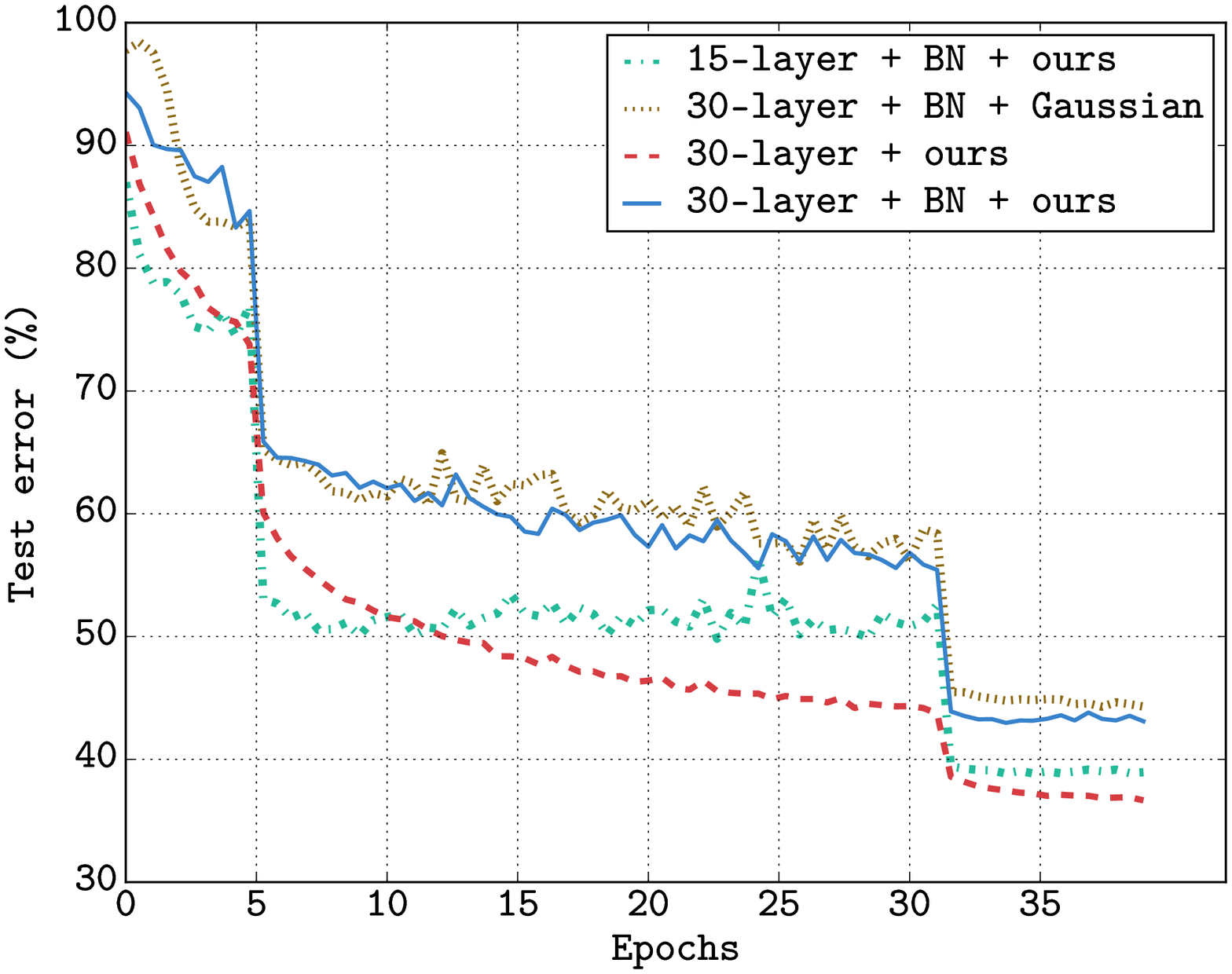}}
     \caption{learning curves of 15/30-layer MPELU networks on ImageNet. (a) training loss: All the 30-layer networks tend to converge. (b) top-1 training error (\%). (c) top-1 test error (\%). the 30-layer networks with BN have higher training/test error than the 15-layer network, suggesting the emergence of the degradation problem \cite{he2015deep}. Somehow surprisingly, if BN is removed, the problem is eliminated (see the red dashed line)}
     \label{fig:degradation phenomenon}
\end{figure}

Firstly, Fig.~\ref{fig:degradation phenomenon}(a) shows the training loss of all the 30-layer networks at the end of training. As we can see, the networks with BN have comparable training loss to the network without BN, demonstrating that they all converge well. Thus, it is most unlikely that the decrease of accuracy is caused by vanishing gradients. Secondly, Fig.~\ref{fig:degradation phenomenon}(b)(c) show the top-1 training/test error rates. Obviously, the 30-layer networks with BN have higher training/test error than the 15-layer counterpart, suggesting the emergence of the degradation problem as described in \cite{he2015deep}. Interestingly, the 30-layer network without BN does not suffer from this problem. It can enjoy the benefit of increasing depth. Note that the only difference among these networks is the usage of BN. Therefore, BN might be an underlying factor causing the degradation problem.

\setlength{\tabcolsep}{4pt}
\begin{table}[t]
\begin{center}
\caption{The statistics (mean and variance) of activations of conv\{1, 7, 14, 20, 27\}. As described in \cite{He_2015_ICCV}, the ReLU network can roughly preserve its variance, which leads to large magnitude of outputs, and thus diverges. As a comparison, the MPELU network can gradually reduce the magnitude, and thus avoid overflow} 
\label{tab:30-layer_convergence}
\scalebox{0.9}{
\begin{tabular}{|c|c|c|c|c|c|c|} \hline
 &   & conv1  & conv7 & conv14 & conv20 & conv27 \\ \hline
\multirow{2}{*}{Mean} & ReLU & 38.95 & 41.25 & 28.37 & 22.52 & 19.61 \\ \cline{2-7}
                      & MPELU & 25.31 & 4.77 & 0.13 & 0.03 & 0.003 \\ \hhline{*7-}
\multirow{2}{*}{Var}  & ReLU & 4196.36 & 4603.98 & 2594.84 & 2381.22 & 2627.62 \\ \cline{2-7}
                      & MPELU & 1840.65 & 74.43 & 0.71 & 0.07 & 0.01 \\ \hline
\end{tabular}}
\end{center}
\end{table}
\setlength{\tabcolsep}{1.4pt}

\noindent \\
\textbf{Comparison of convergence.} Since deeper networks are harder to train, it is good to examine the convergence of activation functions by the 30-layer networks without BN. To this end, four such networks are constructed and initialized from the corresponding method with FAN\_IN, FAN\_OUT, and AVERAGE mode. Experimental results show that the ReLU network fails to converge in all three modes. The PReLU network converges only in the FAN\_OUT mode. On the contrary, ELU/MPELU networks are able to converge in all three modes. These results may be due to the robust to variations of inputs introduced by the left saturation of ELU/MPELU. To verify this, the statistics (mean and variance) are computed. Tab.~\ref{tab:30-layer_convergence} shows that the ReLU network roughly preserves the variance of inputs, which results in very large activations at higher layers and overflow of softmax as discussed in \cite{He_2015_ICCV}. The MPELU network does not suffer from this since it has the left saturation to a small negative value and thereby gradually decreases the variance during forward propagation.

\subsection{Residual Analysis of the Proposed Initialization}

The left side of Eqn.~(\ref{assumption_linearity}) is approximated by the first order Taylor expansion. This section estimates the residual term $R_n(y)$,
\begin{align}
	R_n(y) = \frac{e^{\theta \beta y}}{2!} \alpha (\beta y)^2 \ \ \ (0< \theta <1).
\end{align}
To this end, two cases with and without BN will be considered. 

\noindent \\
\textbf{With BN.} BN are usually adopted immediately before MPELU. Therefore, it is reasonable to assume that the input of MPELU, $y$, has a Gaussian distribution with zero mean at the initialization stage. According to probability theory, over 99.73\% inputs fall into the range of $[-3\sqrt{Var(y)},\ \ 3\sqrt{Var(y)}]$, and in this range only half of them contribute to the residuals. We consider three inputs taking $-\sqrt{Var(y)}$, $-2\sqrt{Var(y)}$, and $-3\sqrt{Var(y)}$ whose corresponding residuals are:
\begin{align}
\label{Eqn: three sigma rule-1}
R_n(\ -3\sqrt{Var(y)}\ ) & = \frac{9e^{-3\theta \beta \sqrt{Var(y)}}}{2} \alpha \beta^2 Var(y) < \frac{9}{2} \alpha \beta^2 Var(y), \\
\label{Eqn: three sigma rule-2}
R_n(\ -2\sqrt{Var(y)}\ ) & = \frac{4e^{-2\theta \beta \sqrt{Var(y)}}}{2} \alpha \beta^2 Var(y) < \frac{4}{2} \alpha \beta^2 Var(y), \\
\label{Eqn: three sigma rule-3}
R_n(\ -\sqrt{Var(y)}\ ) & = \frac{e^{-\theta \beta \sqrt{Var(y)}}}{2} \alpha \beta^2 Var(y)\ \ < \frac{1}{2} \alpha \beta^2 Var(y).
\end{align}
Eqn.~(\ref{Eqn: three sigma rule-1}), (\ref{Eqn: three sigma rule-2}), and (\ref{Eqn: three sigma rule-3}) say that at the initialization, more than 99.865\%, 97.725\%, and 84.135\% (the probability of $y$ falling in [$-3\sqrt{Var(y)}$, $+\infty$], [$-2\sqrt{Var(y)}$, $+\infty$], and [$-\sqrt{Var(y)}$, $+\infty$], respectively) inputs will have the residuals less than $\frac{9}{2} \alpha \beta^2 Var(y)$, $2 \alpha \beta^2 Var(y)$, and $\frac{1}{2} \alpha \beta^2 Var(y)$, respectively. Here, $y$ has unit variance. If $\alpha$ and $\beta$ are initialized with 1, more than 84.135\% inputs will have the residuals less than 0.5. Furthermore, consider some negative input $\hat{y}$ whose residual is less than $10^{-2}$. For $\hat{y}$,
\begin{align}
\label{estimate_residual_y}
	R_n(\hat{y}) &= \frac {e^{\theta \beta \hat y}}{2} \alpha \beta^2 \hat y^2 \ < \  \frac {1}{2} \alpha \beta^2 \hat y^2 \ < \  0.01.
\end{align}
If $\alpha$ and $\beta$ are initialized with 1, then we obtain:
\begin{align}
	\hat y > -\frac {\sqrt{2}}{10 \sqrt{\alpha} \beta} = -0.1414.
\end{align}
This means there will be about 55.57\% inputs having the residuals less than 0.01. Although the residuals are innegligible, Eqn.~(\ref{taylor_result}) still works well in practice. The analysis can be side-verified by Clevert \emph{et al.} work \cite{clevert2015fast} in which they observed that ELU does not show better performance when used with BN. ELU ($\alpha = 1$) behaves more like LReLU ($a = 1$), a linear function, for the whole period of training since most residuals are small, see Tab. \ref{table:4-leaky_relu_experiments}, LReLU (D). 

\noindent \\
\textbf{Without BN.} In this case, it is difficult to estimate the residuals analytically. Fortunately, the residual can be easily computed from the outputs of a convolutional layer. For this purpose, the 30-layer MPELU network without BN from Sec.~\ref{sect:initialization experiments} is adopted. By Eqn.~(\ref{estimate_residual_y}), we consider the inputs of residuals less than \{0.01, 0.5, 2, 4.5\}, or equivalently $\{y\ |\ y > -0.1414\}$, $\{y\ |\ y > -1\}$, $\{y\ |\ y > -2\}$, and $\{y\ |\ y > -3\}$.

\renewcommand{\arraystretch}{0.7}
\setlength{\tabcolsep}{4pt}
\begin{table}[t]
\begin{center}
\caption{The histogram of units for residuals. The bins are (0, 0.01), (0, 0.5), (0, 2), and (0, 4.5). Conv\{1, 7, 14, 20, 27\} are picked from the 27 convolutional layers. For each bin (each row), the deeper the layer, the higher percentage of units fall in it. Once the depth reaches 14, most of units will have residuals 0.5 or less. It is interesting to note that the outputs of the median layer, conv14, approximately have a standard normal distribution }
\label{table:residual_analysis}
\begin{tabular}{|l|l|l|l|l|l|}
\hline
residual & conv1 & conv7 & conv14 & conv20 & conv27 \\
\hline
0.01  & 51.24 & 46.83 & \textbf{56.70} & 78.18  & 89.45 \\
0.5  & 51.65 & 50.00 & \textbf{84.75} & 99.60 & 1 \\
2    & 52.11 & 53.78 & \textbf{97.09} & 1     & 1 \\
4.5  & 52.53 & 57.54 & \textbf{99.71} & 1     & 1 \\
\hline 
\end{tabular}
\end{center}
\end{table}

For simplicity, the statistics are computed every 7 layers. As shown in Tab.~\ref{table:residual_analysis}, the deeper layers have a better approximation for Eqn.~(\ref{assumption_linearity}). Also, once the depth reaches the median, e.g., conv14, most of units will have the residuals less than 0.5. In addition, the statistics of conv14 is very close to a standard normal distribution, which suggests that it plays a role of BN which ensures that gradients can be properly propagated to the lower layers at the initialization. We argue that the residuals are acceptable for the initialization. Sec.~\ref{sect:initialization experiments} has proven the effectiveness of the proposed initialization.

\section{Deep MPELU Residual Networks}
\label{Section-5: Deep MPELU Residual Networks}
Sec.~\ref{Section-4: MPELU_experiments} shows that MPELU and the proposed initialization can bring benefits to the plain networks. This section gives a deep MPELU ResNet to show that the proposed methods are especially suitable for the ResNet architecture \cite{he2015deep} and provides state-of-the-art performance on the CIFAR-10/100 datasets.

\subsection{MPELU and Batch Normalization}
\label{Section: MPELU with BN in ResNet}

This section demonstrates that MPELU, as opposed to ELU, can be used with BN. Clevert \emph{et al.} \cite{clevert2015fast} found that BN can improve ReLU networks, but not (even be harmful to) ELU networks. Observing this, Shah \emph{et al.} \cite{shah2016deep} proposed to remove most BN layers when constructing ResNet using ELU. While removing BN could lower the barrier between them, it tends to diminish the desired regularization properties, which may lead to unexpected negative effect on the generalization capability. We argue that a proper method to alleviate the problem is introducing learnable parameters $\alpha$ and $\beta$.

\renewcommand{\arraystretch}{1}
\setlength{\tabcolsep}{4pt}
\begin{table}[t]
\begin{center}
\caption{
Classification error on CIFAR-10. ReLU is simply replaced with ELU or MPELU. The mean test error over 5 runs is reported except that we show best (mean $\pm$ std) for depth 110. In MPELU ResNet (A), $\alpha$ and $\beta$ are initialized with 1 and updated by SGD with weight decay. For (B), we pay a special attention to MPELU after addition, and initialize $\alpha$ and $\beta$ with 98 and 0.01, respectively}
\label{Table: ReLU-ELU-MPELU-ResNet}
\scalebox{0.9}{
\begin{tabular}{|l|c|c|c|c|c|c|} \hline
\# layers / \# params  & 20  & 32 & 44 & 56 & 110 & \# params \\ \hline
ResNet \cite{he2015deep}& 8.75       & 7.51      & 7.17      & 6.97      & 6.43 (6.61 $\pm$ 0.16) & 1.73M \\ \hline
ELU ResNet         & \textbf{7.980}      & 7.872     & 7.714     & 7.844     & 8.11 (8.36 $\pm$ 0.29) & 1.73M \\ \hline
MPELU ResNet (A)             & 8.12       & 7.35      & 6.90      & 6.72      & 6.21 (6.89 $\pm$ 0.47) & 1.74M \\ \hline
MPELU ResNet (B)             & 8.16 & \textbf{7.12}   & \textbf{6.67} & \textbf{6.27} & \textbf{5.64 (5.77 $\pm$ 0.15)}  & 1.74M \\ \hline
\end{tabular}}
\end{center}
\end{table}
\setlength{\tabcolsep}{1.4pt}

To examine this, we simply replace ReLU with ELU and MPELU in ResNet, keeping any other settings unchanged. $\alpha$ and $\beta$ in MPELU (A) are initialized with 1 and updated by SGD with weight decay. Tab.~\ref{Table: ReLU-ELU-MPELU-ResNet} shows the ELU ResNet performs worse than the original ResNet, potentially demonstrating that BN does not improve the ELU ResNets. On the contrary, the MPELU ResNets (A) consistently reduces the test error for different depths.

The improvement over ELU may receive an explanation from Eqn.~(\ref{MPELU_decompose}) that origins from the learnable parameters in MPELU. Eqn.~(\ref{MPELU_decompose}) suggests that the outputs of BN directly flow into its PReLU submodule and therefore avoid the ELU submodule. Another possible reason comes from the principle of ResNet, a hypothesis that it is easier to optimize the residual mapping than the original mapping. The ResNet architecture is derived from the extreme case of the hypothesis where the identity mapping is optimal. Compared to ReLU and ELU, MPELU covers larger solution space, which allows the solvers to have more opportunities for approximating identity mappings, and therefore improves the performance. To verify this, we pay a special attention to the MPELU layers after addition, where $\alpha$ and $\beta$ are initialized with 98 and 0.01 respectively. By doing so, the shortcut connection and the MPELU layer after addition combine to an identity mapping. Following the philosophy in \cite{he2015deep}, if an identity mapping were optimal, it would be easier to learn an identity mapping by a shortcut connection plus such a MPELU layer than plus a ReLU or ELU layer since neither ReLU or ELU covers the identity mapping. The results are given in MPELU ResNets (B). Tab.~\ref{Table: ReLU-ELU-MPELU-ResNet} shows that MPELU ResNets (B) consistently outperform the counterparts by a large margin, demonstrating the benefit from the larger solution space introduced by the learnable parameters.

\subsection{Network Architectures}
\label{Section: MPELU ResNet Architectures}

\begin{figure}[t]
\centering
\subfloat[][non-bottleneck]{\includegraphics[width=0.25\textwidth]{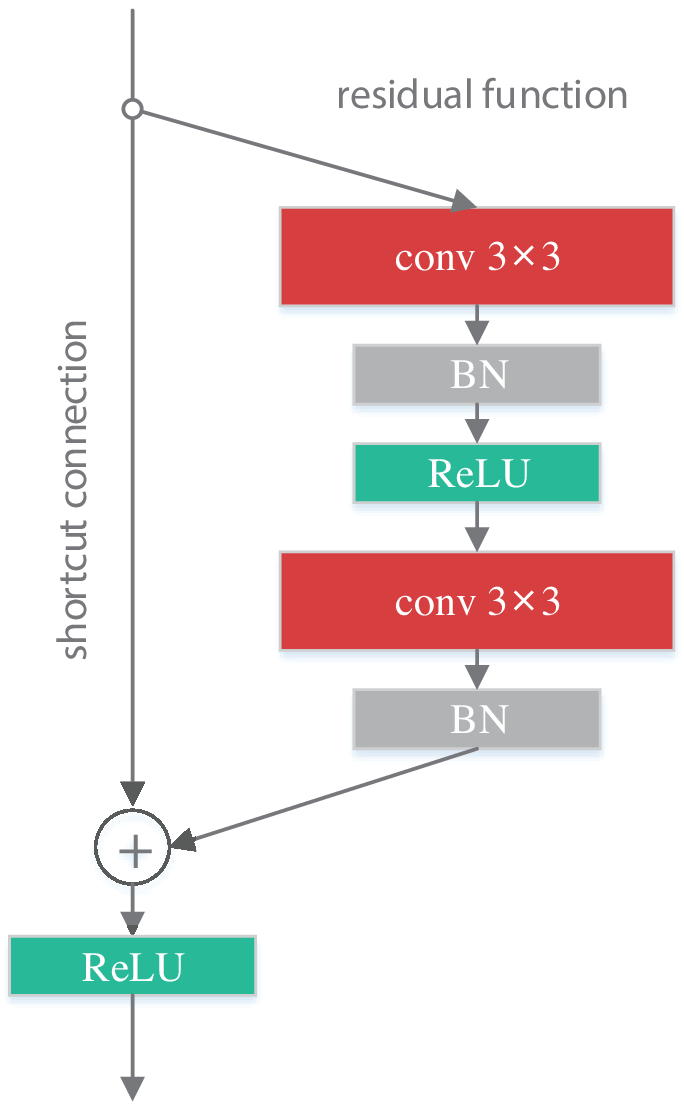}\label{fig: resnet}} ~
\subfloat[][MPELU non-bottleneck]{\includegraphics[width=0.25\textwidth]{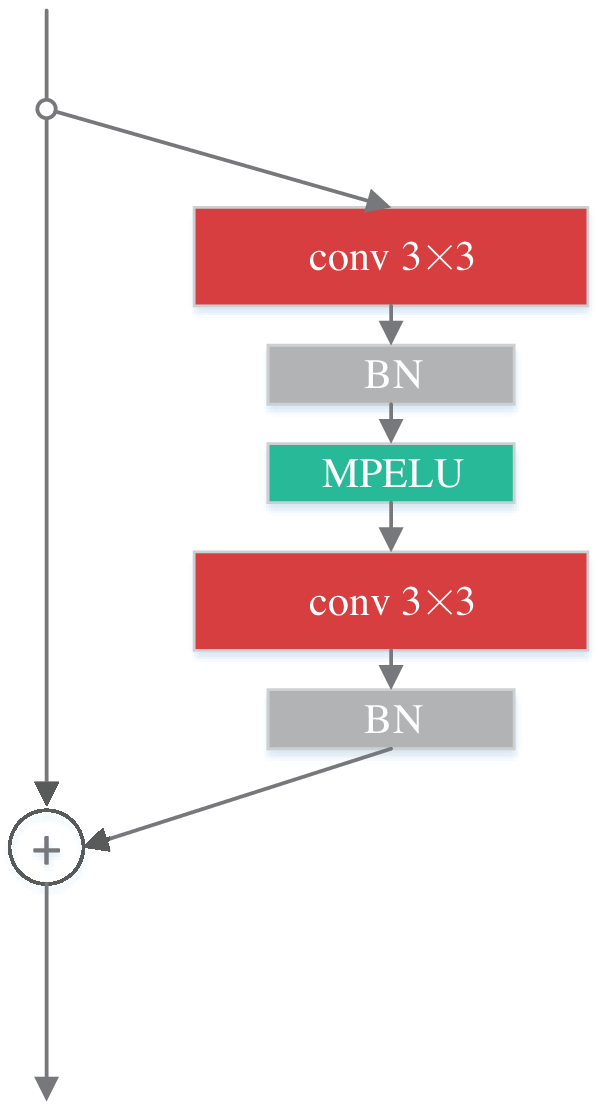}\label{fig: mpelu-resnet}} ~
\subfloat[][full pre-activ. bottleneck]{\includegraphics[width=0.25\textwidth]{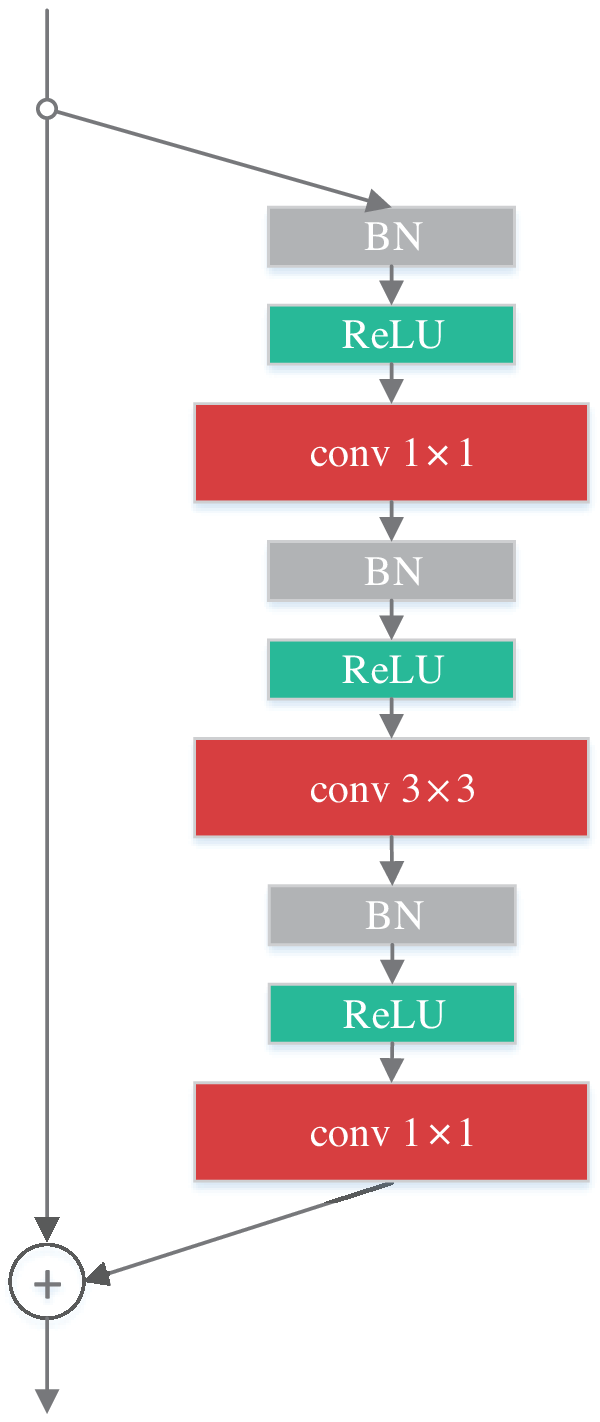}\label{fig: full pre-activation}} ~
\subfloat[][MPELU full pre-activ. bottleneck]{\includegraphics[width=0.25\textwidth]{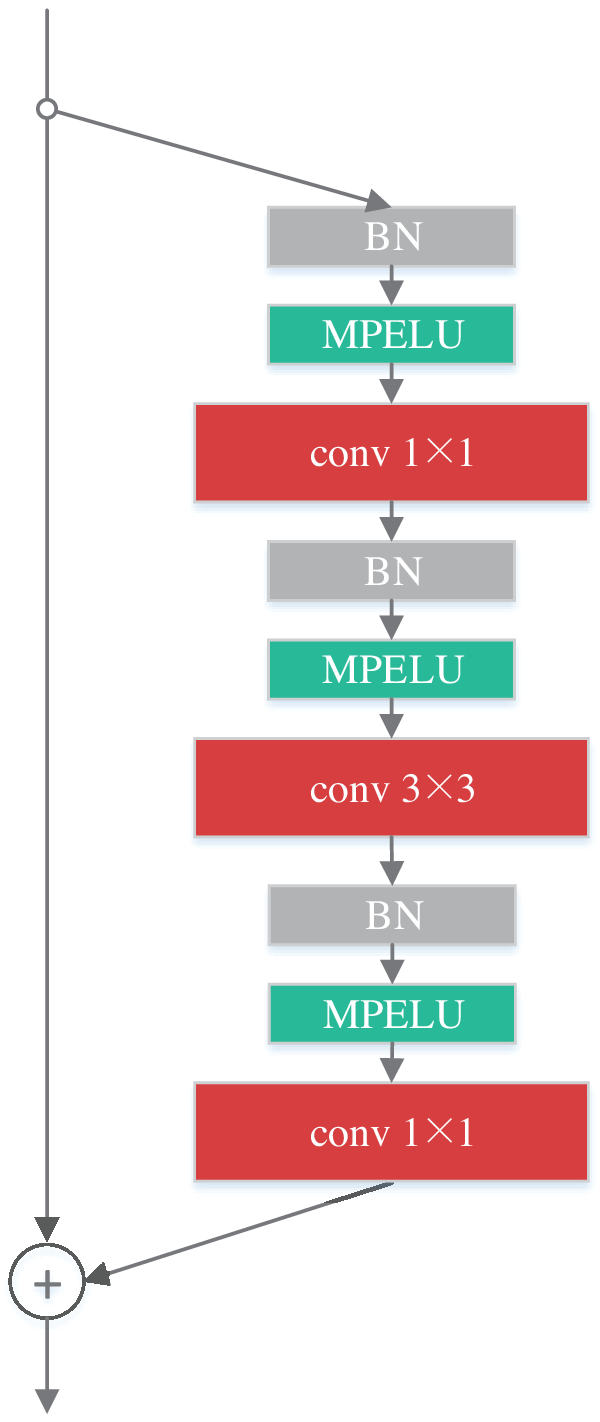}\label{fig: mpelu full pre-activation}}
\caption{Various residual blocks. (a) the non-bottleneck block in \cite{he2015deep}, (b) MPELU non-bottleneck block, (c) the full pre-activation bottleneck block in \cite{He2016}, (d) MPELU full pre-activation bottleneck block}
\label{fig: Various residual blocks}
\end{figure}

He \emph{et al.} \cite{he2015deep,He2016} investigated the usage of activation functions for deep residual networks. The resulted ResNet and Pre-ResNet architectures are highly optimized for ReLU. Even though the performance can be improved by simply replacing ReLU with MPELU as shown in Sec.~\ref{Section: MPELU with BN in ResNet}, we expect that it would benefit from an adjusted deployment. For this reason, this section proposes a variant of the residual architecture, MPELU ResNet which includes two types of blocks, non-bottleneck and bottleneck, as described in the following. \\

\noindent 
\textbf{MPELU Non-bottleneck Residual Block.} This block, (Fig.~\ref{fig: Various residual blocks}(b)), is a simplification of the original non-bottleneck residual block in ResNet \cite{he2015deep} (Fig.~\ref{fig: Various residual blocks}(a)). The experimental results from Sec.~\ref{Section: MPELU with BN in ResNet} suggest that ResNet using MPELU gains more opportunities for finding a better solution than using ReLU or ELU. However, introducing nonlinear units (e.g., MPELU) after addition would still affect the optimization. For example, if an identity mapping were optimal, to the extreme, it would require the solvers to fit an identity mapping by a stack of nonlinear units in addition to pushing the residual functions to zero. Inspired by \cite{He2016,gross2016training}, the identity mapping is directly constructed, as shown in Fig.~\ref{fig: Various residual blocks}(b), by removing the MPELU after addition instead of being fit by the solvers.

\begin{figure}[t]
\centering
\subfloat[][MPELU-only pre-activ. with BN]{\includegraphics[width=0.2\textwidth]{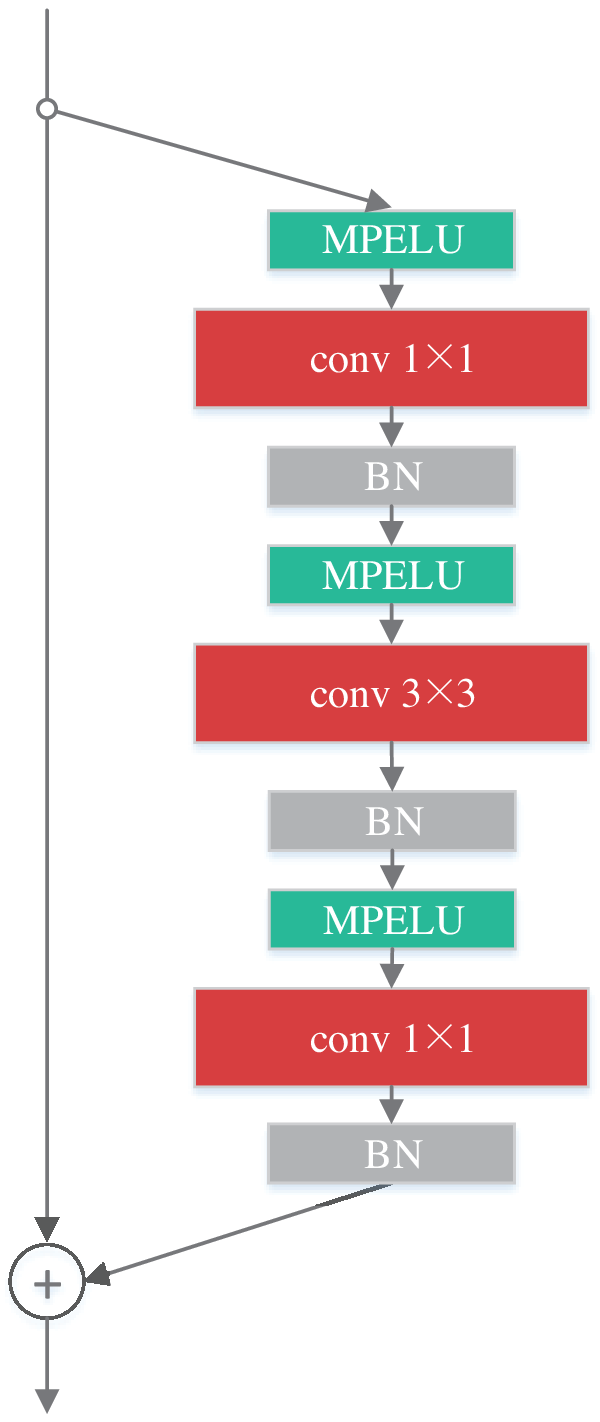}\label{fig: mpelu-halfpre-bn-end}} ~
\subfloat[][MPELU-only pre-activ.]{\includegraphics[width=0.2\textwidth]{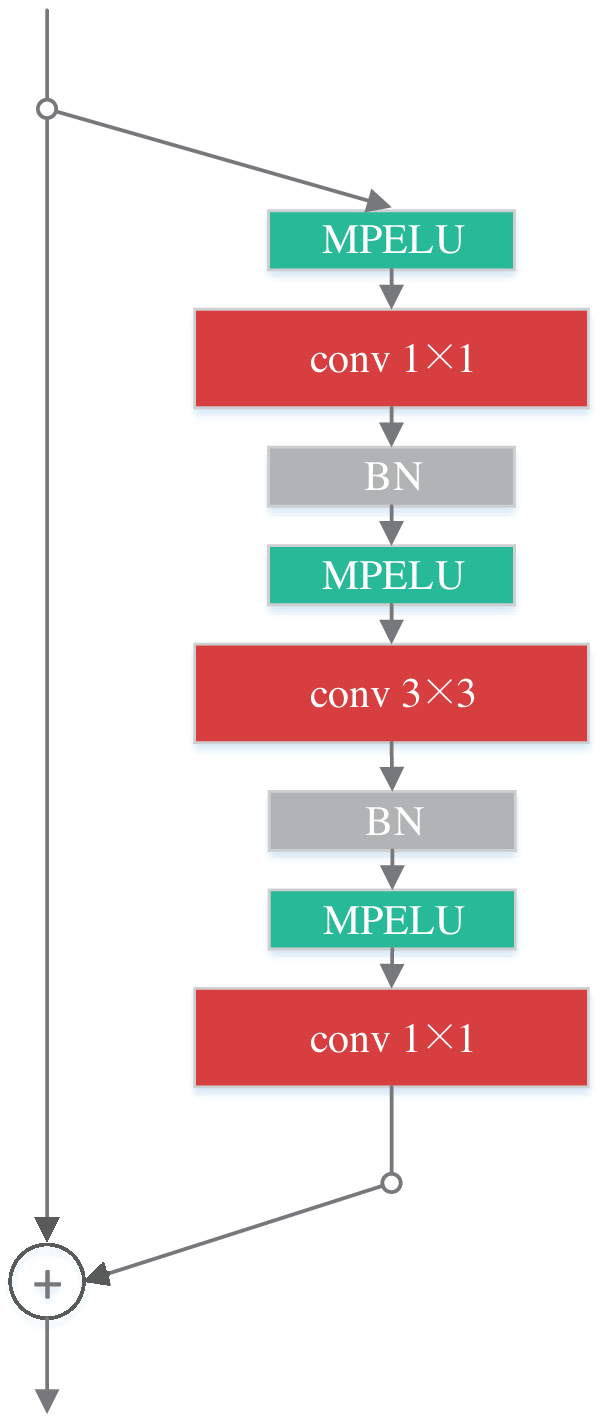}\label{fig: mpelu-only pre-activation}} ~
\subfloat[][nopre with BN before addition]{\includegraphics[width=0.2\textwidth]{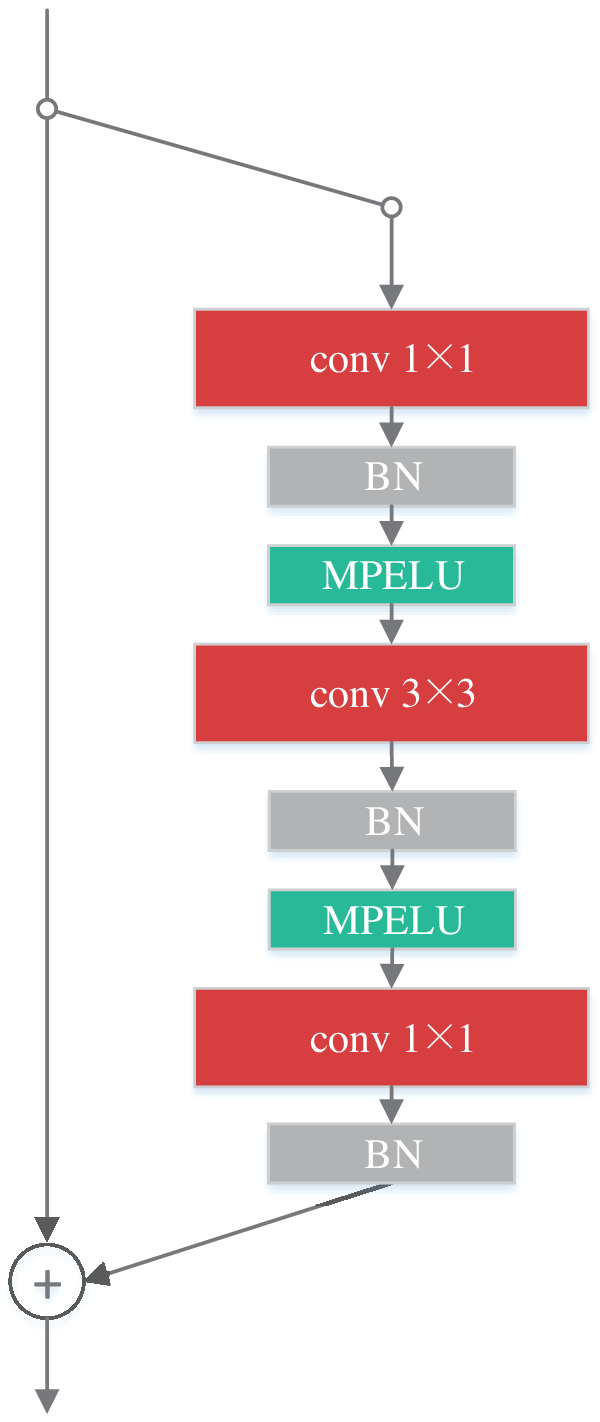}\label{fig: mpelu bn before addition}} ~
\subfloat[][nopre-activ.]{\includegraphics[width=0.2\textwidth]{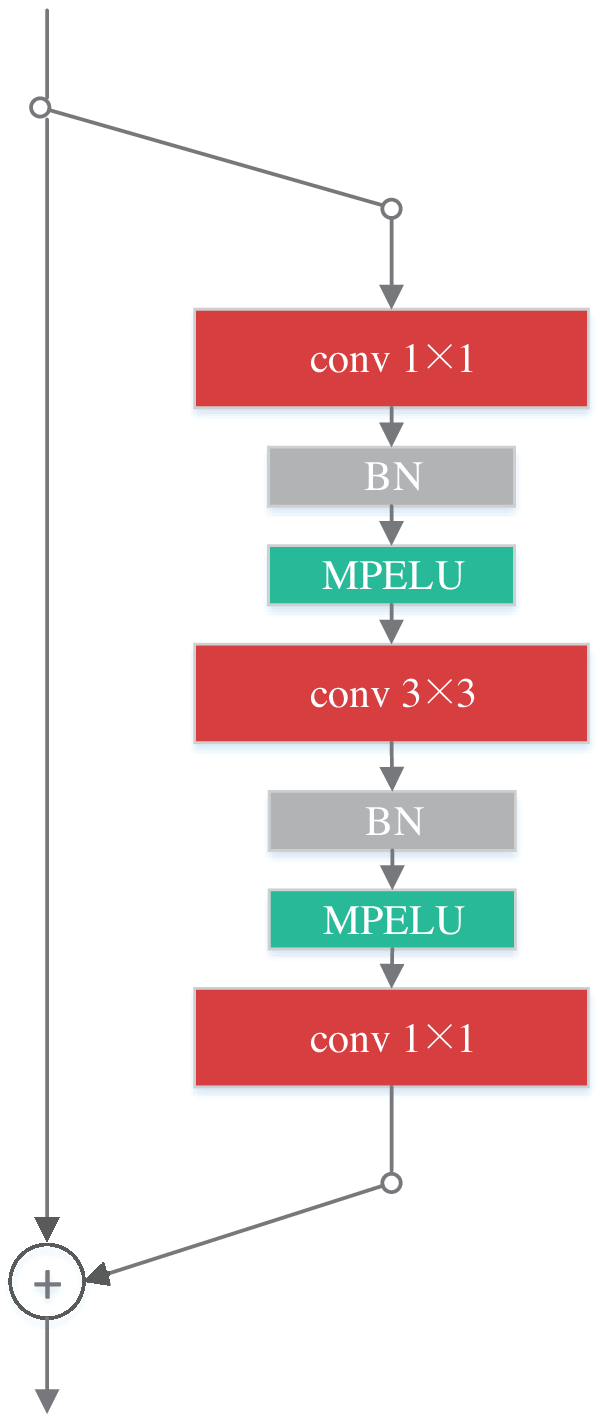}\label{fig: mpelu-nopre}}
\subfloat[][nopre-activ. without BN]{\includegraphics[width=0.2\textwidth]{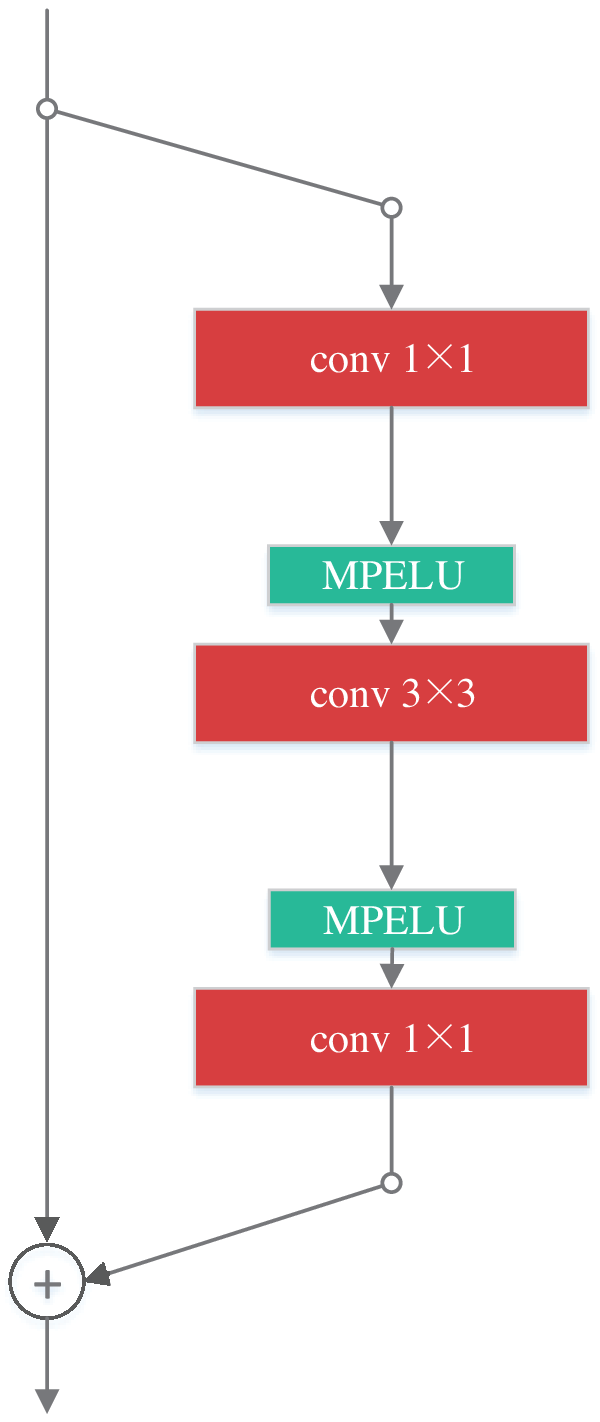}\label{fig: mpelu-nopre-nobn-resnet}}
\caption{Alternatives of residual function. (a) MPELU-only pre-activation block ending with a BN, (b) MPELU-only pre-activation block, (c) nopre-activation with a BN, (d) nopre-activation bottleneck, (e) nopre-activation without BN.}
\label{fig: other nopre alternatives}
\end{figure}

\noindent \\
\textbf{MPELU Bottleneck Residual Block.} A naive MPELU Bottleneck block can be simply obtained by replacing ReLU (Fig.~\ref{fig: Various residual blocks}(c)) with MPELU (Fig.~\ref{fig: Various residual blocks}(d)). This pull pre-activation structure is highly optimized for ReLU. 

This section presents a nopre-activation bottleneck block optimized for MPELU (see Fig.~\ref{fig: other nopre alternatives}(d)). Since the pre-activation part is removed, the complexity and the number of parameters of this block can be largely reduced. As a consequence, the final complexity and the number of parameters of the entire network is comparable to the original. Besides, we adopt a BN (denoted by BN$_1$) plus a MPELU right after the first convolution layer, and a BN (denoted by BN$_{end}$) plus a MPELU right after the last element-wise addition of the entire network. The BN$_1$ and BN$_{end}$ are important for the nopre-activation bottleneck block. We will empirically demonstrate this. In addition to this structure, other alternatives (see Fig.~\ref{fig: other nopre alternatives}) are also investigated. 

\subsection{Results on CIFAR}
\label{Section: MPELU ResNet results}

This section firstly evaluates the variants and alternatives of the proposed MPELU ResNet, then compares it to the state-of-the-art architectures. The implementation details are given in appendix. 

\setlength{\tabcolsep}{4pt}
\begin{table}[t]
\begin{center}
\caption{Test error (\%) of non-bottleneck architectures on CIFAR-10. We try different learning rate and weight decay multipliers for $\alpha$ and $\beta$, and pick the one that gets the best performance. We retrained the original ResNet for 200 epochs and denote the results by *} 
\label{Table: MPELU non-bottleneck on cifar10}
\scalebox{0.9}{
\hspace*{-22pt}
\begin{tabular}{|l|c|c|c|c|c|c|c|} \hline
Fig. / \# layers / \# params  & Fig. & 20  & 32 & 44 & 56 & 110 & \# params \\ \hline
ResNet \cite{he2015deep} & Fig.~\ref{fig: Various residual blocks}(a) & 8.75       & 7.51      & 7.17      & 6.97      & 6.43 (6.61 $\pm$ 0.16) & 1.73M \\ \hline
ResNet \cite{he2015deep}* & Fig.~\ref{fig: Various residual blocks}(a) & 8.16       & 7.06      & 6.99      & 6.58      & 6.27 (6.40 $\pm$ 0.18) & 1.73M \\ \hline
MPELU ResNet (non-bottle.) & Fig.~\ref{fig: Various residual blocks}(b) & \textbf{7.71} & \textbf{6.73}   & \textbf{6.26} & \textbf{5.95} & \textbf{5.35 (5.47 $\pm$ 0.14)}  & 1.74M \\ \hline
\end{tabular}}
\end{center}
\end{table}
\setlength{\tabcolsep}{1.4pt}

\setlength{\tabcolsep}{4pt}
\begin{table}[t]
\begin{center}
\caption{Test error (\%) of bottleneck architectures on CIFAR-10. $\alpha$ and $\beta$ are initialized with 0.25 and 1, respectively, and updated by SGD with weight decay}
\label{Table: MPELU bottleneck variants on cifar10}
\scalebox{0.9}{
\begin{tabular}{|l|c|c|c|c|c|c|c|} \hline
Fig. / \# layers / \# params   & Fig. & 164 & \# params \\ \hline
the original Pre-ResNet \cite{He2016} & Fig.~\ref{fig: Various residual blocks}(c) & 5.46 & 1.703M \\ \hline
MPELU full pre-activ. & Fig.~\ref{fig: Various residual blocks}(d) & 5.20 (5.32 $\pm$ 0.13) & 1.728M \\ \hline
MPELU-only pre-activ. with BN & Fig.~\ref{fig: other nopre alternatives}(a) & diverged within few steps & 1.727M \\ \hline
MPELU-only pre-activ. & Fig.~\ref{fig: other nopre alternatives}(b) & 5.49 & 1.712M \\ \hline
MPELU nopre with BN & Fig.~\ref{fig: other nopre alternatives}(c) & diverged within few steps & 1.713M \\ \hline
MPELU nopre & Fig.~\ref{fig: other nopre alternatives}(d) & \textbf{4.87 (5.04 $\pm$ 0.14)} & 1.696M \\ \hline
MPELU nopre (no BN$_1$ and BN$_{end}$)& - & diverged within few steps & 1.696M \\ \hline
MPELU nopre (no BN$_{1}$)& - & 5.29 & 1.696M \\ \hline
MPELU nopre without BN & Fig.~\ref{fig: other nopre alternatives}(e) & diverged within few steps & 1.688M \\ \hline
\end{tabular}}
\end{center}
\end{table}
\setlength{\tabcolsep}{1.4pt}

\noindent \\
\textbf{Classification Results.} For shallower architectures, the MPELU ResNets (non-bottle.) are considered. Tab.~\ref{Table: MPELU non-bottleneck on cifar10} shows that the MPELU ResNets (non-bottle.) achieve consistent improvement with negligible increase of parameters. For example, the 110-layer MPELU ResNet reduces the mean test error rate to 5.47\%, which is 1.14\% lower than the original ResNet-110. Note that this improvement is obtained merely via a simple strategy -- changing the usage of activation functions, demonstrating the benefit from MPELU.

When the networks go deeper (164 layers), we focus on the bottleneck architectures to reduce the time/memory complexity as done in \cite{he2015deep}. Tab.~\ref{Table: MPELU bottleneck variants on cifar10} shows that the MPELU full pre-activ., Fig.~\ref{fig: Various residual blocks}(d), provides a marginal decrease in the mean test error rate from 5.46\% to 5.32\% compared to the original Pre-ResNet, Fig.~\ref{fig: Various residual blocks}(c). This is done by simply replacing ReLU with MPELU. For the MPELU-only pre-activ. with BN (Fig.~\ref{fig: other nopre alternatives}(a)), the network fails to converge under the initial learning rate 0.1. Following \cite{he2015deep}, we warm up the training using learning rate 0.01 for one epoch, then switch back to 0.1. With this policy, the network is able to converge but to a worse solution than the full pre-activ. architecture. Based on this observation, we keep the pre-activation part and remove the BN before addition (see Fig.~\ref{fig: other nopre alternatives}(b)). Interestingly, the network can converge without warming up, leading to the mean test error 5.49\% which is also worse than the full pre-activ. architecture. Through these results, the MPELU-only pre-activ. architectures are not considered in the rest of the paper.

We focus on the MPELU nopre architecture (Fig.~\ref{fig: other nopre alternatives}(d)), and its variants. Somehow surprisingly, as shown in Tab.~\ref{Table: MPELU bottleneck variants on cifar10}, simply removing the pre-activation brings about lower test error rate with less parameters and complexity, which suggests that the deep residual architectures have the potential to enjoy the benefit from MPELU. In addition, the performance is also examined by adding more BN layers to and removing BN layers from the MPELU nopre architecture. For the former case (Fig.~\ref{fig: other nopre alternatives}(c)), as demonstrated in Tab.~\ref{Table: MPELU bottleneck variants on cifar10}, adding one more BN before addition makes the network diverge within few steps. Seeing this, we tried the warming up and found that the network converged well. Combining this phenomenon with the observations of Fig.~\ref{fig: other nopre alternatives}(a) and ResNet-110 \cite{he2015deep}, we suspect that the BN before addition would exert a negative impact on the gradient signals so that we have to lower the initial learning rate to warm up the training. For the latter case, removing all the BN from the residual function (see Fig.~\ref{fig: other nopre alternatives}(e)) also leads to divergence. Again, the same result happens when BN$_1$ and BN$_{end}$ are removed from the MPELU nopre. However, if keeping BN$_{end}$, the network still converges and performs slightly worse (5.29\% $vs.$ 5.04\% mean test error). These results suggest that BN$_1$ and BN$_{end}$ are important to the nopre architecture.

Considering the time/memory complexity and model size, the MPELU nopre is picked as the proposed bottleneck architecture of this paper and used to compared to other state-of-the-art methods.

\renewcommand{\arraystretch}{0.7}
\setlength{\tabcolsep}{4pt}
\begin{table}[t]
\begin{center}
\caption{Comparison to state-of-the-art methods on CIFAR-10/100. MPELU are initialized with $\alpha = 0.25\ or\ 1$ and $\beta = 1$ that are updated by SGD with weight decay. $\dagger$ denotes that the hyper-parameter settings follow \cite{huang2016densely} (see appendix). Our results are based on the best of 5 runs with mean $\pm$ std}
\label{Table: compared to state-of-the-arts}
\scalebox{0.9}{
\begin{tabular}{l|l|c|c|c|c} \hline 
Method   & settings & depth & \# params & CIFAR-10 & CIFAR-100 \\ \hline

NIN \cite{2013arXiv1312.4400L} & - & - & - & 8.81 & - \\
DSN \cite{lee2015deeply} & - & - & - & 7.97 & 34.57 \\
All-CNN \cite{springenberg2014striving} & - & - & - & 7.25 & 33.71 \\
Highway \cite{NIPS2015_5850} & - & - & - & 7.72 & 32.39 \\
ELU \cite{clevert2015fast} & - & - & - & 6.55 & 24.28 \\
Fitnets \cite{romero2014fitnets} & - & - & - & 8.39 & 35.04 \\ \hline
\multirow{2}{*}{ResNet \cite{he2015deep}} & - & 110 & 1.7M & 6.61 & - \\
                                           & - & 1202 & 19.4M & 7.93 & - \\ \hline
\multirow{2}{*}{sto. ResNet \cite{huang2016deep}} & - & 110 & 1.7M & 5.23 & 24.58 \\
                                                   & - & 1202 & 10.2M & 4.91 & - \\ \hline
\multirow{2}{*}{Wide ResNet \cite{zagoruyko2016wide}} & k = 8 & 16 & 11.0M & 4.81 & 22.07 \\
                                                      & k = 10 & 28 & 36.5M & 4.17 & 20.50 \\ \hline
\multirow{2}{*}{Pre-ResNet \cite{He2016}} & - & 164 & 1.7M & 5.46 & 24.33 \\
                 & - & 1001 & 10.2M & 4.62 (4.69 $\pm$ 0.20) & 22.71 (22.68 $\pm$ 0.22) \\ \hline 
  & $\alpha = 1$ & 164$^\dagger$ & 1.696M & 4.58 (4.67 $\pm$ 0.06) & 21.35 (21.78 $\pm$ 0.33)  \\
MPELU nopre& $\beta = 1$ & 1001$^\dagger$ & 10.28M & \textbf{3.63 (3.78 $\pm$ 0.09)} & \textbf{18.96 (19.08 $\pm$ 0.16)}\\ \cline{2-6}
ResNet& $\alpha = 0.25$ & 164 & 1.696M & 4.87 (5.06 $\pm$ 0.14) & 23.16 (23.29 $\pm$ 0.11)  \\
(Fig. 5(d))& $\beta = 1$ & 164$^\dagger$ & 1.696M & 4.43 (4.53 $\pm$ 0.12) & 21.69 (21.88 $\pm$ 0.19) \\
& & 1001$^\dagger$ & 10.28M & \textbf{\color{blue}{3.57 (3.71 $\pm$ 0.11)}} & \textbf{\color{blue}{18.81 (18.98 $\pm$ 0.19)}} \\ \hline
\end{tabular}}
\end{center}
\end{table}
\setlength{\tabcolsep}{1.4pt}

\noindent \\
\textbf{Comparison to state-of-the-art methods.} To compare to the state-of-the-art methods, we adopt an aggressive training strategy from \cite{huang2016densely} (See appendix for details), denoted by the symbol $\dagger$. 

The test error rate is given in Tab.~\ref{Table: compared to state-of-the-arts}. It is easy to see that with the training strategy $\dagger$, the mean test error of MPELU nopre ResNet-164 ($\alpha = 0.25$) is considerably reduced especially on CIFAR-100 dataset (21.88\% $vs.$ 23.29\%). This might be because that CIFAR-100 is challenger than CIFAR-10. Training for more epochs with large learning rate would help the model learn the underlying elusive concepts. Interestingly, changing the initial value of $\alpha$ to 1 in MPELU can further improve the test error on CIFAR-100 (21.78\%) but not on CIFAR-10 (4.67\%). For comparison, we also trained the 1001-layer MPELU nopre ResNet. Tab.~\ref{Table: compared to state-of-the-arts} shows that even though more parameters are introduced, the MPELU ResNet architectures do not suffer from overfitting and still enjoy the performance gains from increased parameters and depth. The best results from the proposed MPELU nopre ResNet-1001 are 3.57\% test error on CIFAR-10 and 18.81\% on CIFAR-100, which are considerably lower than those by the original Pre-ResNet \cite{He2016}.

\section{Conclusions}
\label{Section-6: conclusion}

Activation function is the pivotal component of deep neural networks. Recently, some work on this subject has been proposed. This paper generalized the existing work to a new Multiple Parametric Exponential Linear Units (MPELU). By introducing the learnable parameters, MPELU can become the rectified or the exponential linear units and combine their advantages. Comprehensive experiments via networks of varying depth (from 9-layer NIN \cite{2013arXiv1312.4400L} to 1001-layer ResNet \cite{he2015deep}) are conducted to examine the performance of MPELU. Experimental results showed that MPELU can bring benefits to the classification performance and the convergence of deep networks. In addition, MPELU can work with Batch Normalization as opposed to ELU. Weight initialization is also an important factor in deep neural networks. This paper proposes an initialization for networks using exponential linear units, which complements the current theory of this field. To our knowledge, this is the first method that gives an analytic solution for networks using exponential linear units. Experimental results demonstrated that the proposed initialization not only enable the training of very deep networks using exponential linear units, but leads to better generalization performance. In addition, these experiments suggested that Batch Normalization might be one of factors that caused the degradation problem. Finally, this paper investigated the usage of MPELU with ResNet and presented a deep MPELU residual networks which achieved state-of-the-art accuracy on the CIFAR-10/100 datasets.

\section*{Acknowledgement}
We would like to acknowledge NVIDIA Corporation for donating the Titan X GPU and supporting this research. This work was supported by the National Natural Science Foundation of China 
(Grants No., NSFC-61402046, NSFC-61471067, NSFC-81671651), 
Fund for Beijing University of Posts and 
Telecommunications (Grants No., 2013XD-04, 2015XD-02), 
Fund for National Great Science Specific Project (Grants No. 2014ZX03002002-004), 
Fund for Beijing Key Laboratory of Work Safety and Intelligent Monitoring. 

\section*{Appendix: Implementation Details}

\noindent
\textbf{NIN on CIFAR-10 (Sec.~\ref{Section: experiments on cifar10}).} During training, all the models are trained using SGD with batch size 128 for 120k iterations (around 307 epochs). The learning rate is initially set to 0.1, and then decreased by a factor of 10 after 100k iterations. The weight decay and momentum are 0.0001 and 0.9. The weights are initialized from a zero-mean Gaussian distribution with 0.01 standard deviation. $\alpha$ and $\beta$ in MPELU are initialized with 0.25 or 1, and updated by SGD without weight decay. During test, we adopt the single-view test. Following \cite{goodfellow2013maxout,2013arXiv1312.4400L,NIPS2015_5850}, the data is preprocessed with global contrast normalization and ZCA whitening. When data augmentation is used, the $28\times28$ patches are randomly cropped from the preprocessed images, and then flipped with a probability of 50\%.

\noindent
\textbf{The 15-layer networks on ImageNet (Sec.~\ref{Section: experiments on ImagenNet}).} The models are trained by SGD with mini-batch size of 64 for 750k iterations (37.5 epochs). The learning rate is 0.01 initially, then divided by 10 at 100k and 600k iterations. The weight decay and momentum are 0.0005 and 0.9, respectively. All of images are scaled to $256\times256$ pixels. During training, a $224\times224$ sub image is randomly sampled from the original image or its flipped version. No further data augmentation is used. During test, we adopt the single-view test.

\noindent
\textbf{MPELU ResNet on CIFAR-10/100 (Sec.~\ref{Section: MPELU ResNet results}).} The implementation details mainly follow \cite{he2015deep} and the fb.resnet.torch \cite{gross2016training}. Specifically, the models are trained by SGD with batch size of 128 for 200 epochs (no warming up). The learning rate is initially set to 0.1, then decreased by a factor of 10 at 81 and 122 epochs. The weight decay is set to 0.0001, and the momentum is set to 0.9. MPELU are initialized with $\alpha = 0.25\ or\ 1$ and $\beta = 1$ that are updated by SGD with weight decay. All the MPELU models are initialized from the proposed method (Sec.~\ref{Section: weight initialization}). For comparison, we follow the standard data augmentation implemented by fb.resnet.torch \cite{gross2016training}: each image is padded with 4 pixels and then a 32$\times$32 patch is randomly cropped from it or its horizontal flip version. When the aggressive training strategy $\dagger$ from \cite{huang2016densely} is adopted, the models are trained for 300 epochs. The batch size is 64 on two Titan X GPUs (32 each). The learning rate is initially at 0.1, then decreased by a factor of 10 at 150 and 225 epochs.

\bibliographystyle{splncs}
\bibliography{mpeluv3}
\end{document}